\documentclass[11pt]{article}
\usepackage{amsmath}

\usepackage[final]{acl}
\usepackage{subcaption}
\usepackage{booktabs}
\usepackage{times}
\usepackage{amssymb}
\usepackage{latexsym}
\usepackage{hyperref}
\usepackage{tcolorbox}
\usepackage{float}
\usepackage{multirow}

\usepackage[T1]{fontenc}
\usepackage[utf8]{inputenc}

\usepackage{microtype}
\usepackage{inconsolata}

\usepackage{graphicx}
\usepackage{tcolorbox}
\title{Is Convergence Inevitable?\\Tracing Output Homogeneity Back to Base Models}

\definecolor{red_orange}{HTML}{ed5655}
\definecolor{pink}{HTML}{ff55be}
\author{Alexandrine Fortier}

\author{
 \textbf{Alexandrine Fortier},
 \textbf{Hazel Chen},
 \textbf{Peter West}
\\
 University of British Columbia
\\
 \small{
    {alexf01@cs.ubc.ca}
 }
}

\begin{document}
\maketitle
\begin{abstract}

% The training objective of language models (LMs)---optimizing toward a single ideal output---goes against the core idea of creativity. The lack of diversity in LMs is known to be linked to the alignment process, but causal study of its role is missing. In this work, we propose a controlled investigation of how each stage of the training pipeline contributes to output convergence, using metaphor generation as a probe. We find that convergence strongly emerges during the instruction-tuning stage (SFT) and is reinforced throughout the following alignment stages. To understand why SFT causes convergence, we design and conduct a set of controlled SFT trainings, curating the fine-tuning data to control output convergence. We find that SFT enforces convergence by imposing an output structure, and that ideas introduced in the SFT training set can strongly amplify it. Beyond alignment, we find that converging outputs cannot be predicted from the pretraining corpus, as the output distribution is not representative of the corpus. Altogether, our findings confirm the need for pluralistic alignment methods.

The lack of diversity in LM content is widely attributed to the alignment process, but how and where exactly in the pipeline this collapse begins is unknown. We argue that output homogeneity is likely learned during the pretraining phase, and only \emph{revealed} or magnified during the alignment process. Specifically, we find that semantic convergence is observed from the first alignment stage---the instruction-tuning phase (SFT)---suggesting that homogeneity might already exist in the pre-alignment model. To investigate this, we conduct controlled SFT experiments examining how training data influences output convergence on specific input/output pairs. We find that convergence can be revealed and amplified, but not introduced by the SFT data, supporting its role as a catalyst rather than a cause. To further test whether homogeneity originates before alignment, we measure convergence in base models. We find that instruct-like collapse can be induced through prompting alone, even without alignment. Taken together, our results suggest that semantic convergence may arise naturally from the objectives underlying LM training, making it difficult to mitigate through post-alignment interventions alone.

\end{abstract}

\section{Introduction}
% converge onto a narrow output space, producing the \textit{ideal}, but expected answer.

% Language models (LMs) often converge onto a narrow output space, producing repetitive and expected content, but the cause of this collapse is vague. 

% Language models (LMs) are known to produce repetitive outputs lacking diversity, yet where and how this collapse arises in the training pipeline is unknown. 

\begin{figure*}
    \centering
    \includegraphics[width=1\linewidth]{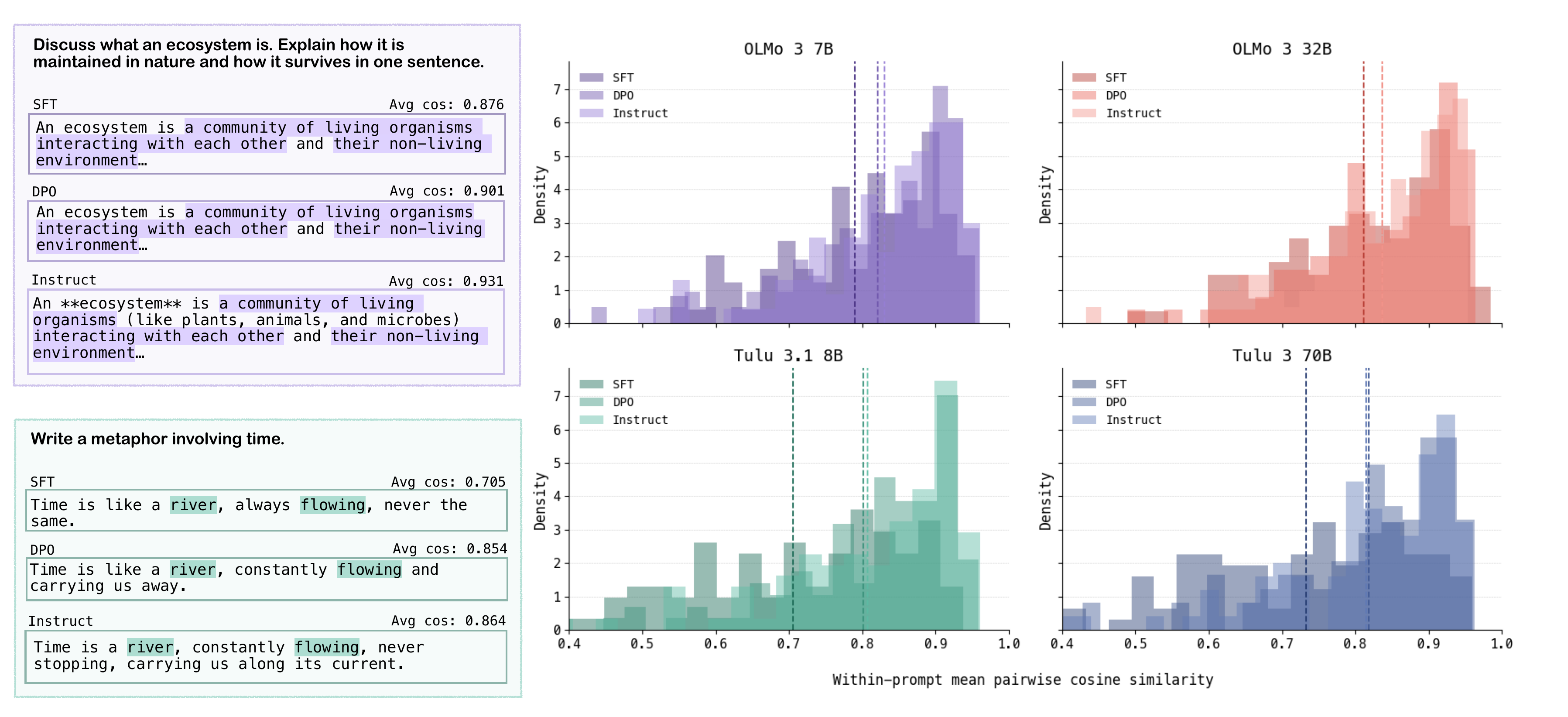}
    \caption{Distributions of mean pairwise cosine similarity across all 100 prompts for OLMo 3 (7B, 32B) and Tulu 3 (8B, 70B) show that \textbf{convergence is already pronounced after SFT}, with distributions right-shifted and sharing overlap with DPO and Instruct. Qualitative examples further illustrate this: sample responses from SFT, DPO, and Instruct checkpoints of OLMo-3-7B (purple) and Tulu 3.1 8B (green) read as paraphrases of one another. Dashed lines indicate the mean cosine similarity per stage.}
    \label{fig:alignment_stages}
\end{figure*}

Language models (LMs) often converge onto a narrow output space, producing repetitive and expected content, but it remains unclear whether this semantic collapse is an inherent consequence of pretraining or a behavior introduced by post-training alignment. As humans increasingly interact with LMs for open-ended tasks, from creative writing \cite{Kumar_2025, Anderson_2024} to soliciting life advice \cite{10.1145/3715275.3732039, stade2024large}, understanding the underlying cause of this homogeneity becomes a societal issue that goes beyond making systems efficient. Previous work found that the alignment process can reduce creativity and randomness across a range of tasks, such as writing and summarization \cite{west2025base, kirk2024understandingeffectsrlhfllm}, word-color associations \cite{Murthy_2025}, as well as world modeling \cite{li2024predictingvsactingtradeoff}. We instead approach this issue from another angle: does alignment induce convergence, or simply reveal patterns learned during pretraining?

% convergence might have been there all along and only further revealed through alignment. 

% We argue that convergence might have been there all along

% We argue that convergence might have been there all along

% Yet, these studies remain heavily observational and lack causal analysis; the path to homogeneity remains untraced.

% In this work, we investigate how and where output homogeneity emerges in the LM training pipeline. We argue that highly converging patterns are likely learned during the pretraining phase, and only revealed or magnified during the alignment process. 

First, we examine semantic similarity on a set of open-ended answers, across the standard alignment stages: supervised fine-tuning (SFT), direct preference optimization (DPO), and reinforcement learning with verifiable rewards (RLVR). Remarkably, we find that output convergence already strongly manifests after the first alignment stage, the instruction-tuning phase---SFT. This suggests that convergence is either a product of SFT or already present in the pre-alignment model, and only revealed through instruction-tuning. 

From this, we conduct controlled SFT experiments examining how training data influences output convergence on specific input/output pairs. We show that simply enforcing an answer format is sufficient to \emph{reveal} latent convergence. To test whether convergence can be shifted by the SFT data, we further train on datasets containing both highly convergent metaphors and deliberately implausible metaphors. We find that models only reinforce patterns consistent with their pre-existing distribution, while failing to absorb out-of-distribution patterns. We conclude that convergence can be amplified, but not introduced. 

To confirm our hypothesis that homogeneity is latent, we directly evaluate diversity in base models. We find that base models, under specific prompting conditions, elicit instruct-like behavior and converge toward preferred outputs.

% few-shot examples or an assistant-style persona are sufficient to elicit instruct-like behavior, leading base models to converge toward preferred outputs, similarly to aligned models.

Overall, our work suggests that converging patterns are latent in pre-alignment models and become revealed and amplified during instruction tuning. This points to semantic convergence as a property of pretraining, and implies that post-alignment interventions alone may be insufficient to mitigate it.

\section{Convergence Through Alignment Stages}
\label{sec: convergence_stages}

Aligned models have been shown to produce less diverse and creative outputs than base models \citep{west2025base, Murthy_2025}. Yet it remains unclear whether convergence originates in preference optimization (DPO, RL post-training) or arises earlier during instruction-tuning. Using open-ended queries, we evaluate output homogeneity at different stages of the alignment process---SFT, DPO, and RLVR---in order to identify where convergence begins. Broadly, we find that outputs are already strikingly similar after the first alignment stage, suggesting convergence may precede alignment entirely.\\

% Among the components of alignment, which is responsible for collapsing output diversity?

% We find that the average pairwise semantic similarity of responses after model experience SFT is already high up in the 0.7-0.8 range of all models, hinting that homogeneity already exists strongly after the first alignment stage (SFT), only then reinforced throughout the following stages. Subsequent stages (DPO, RLVR) shift the distribution modestly further, but the predominant convergence is established at SFT. Our results shows that convergence happens before any reward models/reward signals are introduced. This finding shows something counterintuitive: SFT is understood to enable instruction following without introducing new knowledge or response preferences \citep{zhou2023limaalignment}, yet it is the stage at which outputs collapse. \\

\textbf{Alignment stages.}
Through the alignment process, the model is shaped to follow instructions and agree with human needs and values. Although alignment pipelines can be complex and comprise various intricate steps, we here focus on the three common stages: SFT, DPO, and RLVR. First, the model goes through instruction-tuning via SFT \cite{ouyang2022traininglanguagemodelsfollow}. During the SFT stage, the model is fine-tuned on data structured in an instruction and answer template. The model learns how to follow instructions, but also how to structure its answers efficiently. The next step is the DPO, where the model is trained on ranked answers \cite{rafailov2024directpreferenceoptimizationlanguage}. Finally, the model is further refined through reinforcement learning (RL), where it is rewarded for producing correct outputs \cite{lambert2025tulu3pushingfrontiers, ouyang2022traininglanguagemodelsfollow}. In the Tülu 3 and Olmo 3 pipelines, reinforcement learning with verifiable rewards (RLVR) is used. This produces the final \textit{Instruct} model (\autoref{fig:alignment_stages}). We evaluate output convergence after each main alignment step: SFT, DPO, and RLVR. 

\subsection{Experiment}

\textbf{Setup.}
We conduct our evaluation on the dataset \textsc{Infinity-Chat100} \cite{jiang2025artificial}. This set comprises 100 open-ended user queries, known to produce converging outputs, intra- and inter-model, despite admitting a wide range of correct answers. For each model and alignment stage, we sample 50 responses per query, using top-p=0.9
and temperature=1.0, consistent with \citet{jiang2025artificial}. Outputs are embedded using OpenAI's \textit{text-embedding-3-small}. We measure output uniformness using the mean pairwise cosine similarity of a model's responses for a given prompt. 
More details about inference parameters can be found in \autoref{inf_details}. \\

% For each specific alignment stage of the model of interest, we embed the 50 responses using OpenAI's \textit{text-embedding-3-small} model and compute cosine similarities \citep{openai2024textembedding3small}. Following the metric and decoding setting from \citet{jiang2025artificial}, We average cosine similarities into a uniformness score. Higher values indicate more semantically similar — and thus more uniform — outputs. 

\textbf{Models.}  
We evaluate the behavior of the Tülu 3 suite \cite{lambert2025tulu3pushingfrontiers} at 8B and 70B parameters, and on the OLMo 3 suite \cite{olmo2026olmo3} at 7B and 32B parameters. The Tülu-3 suite is built on the Llama-3.1 base model \cite{grattafiori2024llama3herdmodels}, and the OLMo 3 suite is pretrained from scratch. Since extracting coherent responses from base models requires task-specific prompting that would confound direct comparison with instruction-following checkpoints, this section focuses on post-training stages. We examine base model convergence separately in \autoref{converg_base}.\\

\subsection{Results}
% Our results show that SFT introduce uniformness of responses, DPO and RLVR add a bit but not a lot more convergence. This should be pretty self-explanatory from the figure. We see that after the models go through SFT, they are already producing responses that are highly similar to each other. 
Convergence is already established after SFT. Figure~\ref{fig:alignment_stages} shows the distribution of mean pairwise cosine similarity across 100 prompts at each alignment stage. Across all models, the distribution is heavily right-shifted after SFT, with the bulk of queries concentrated above 0.7 in mean pairwise cosine similarity. DPO and RLVR only shift the mean modestly further rightward. In addition, we show in the same figure that responses sampled from the same post-SFT model to the same prompt read as paraphrases of one another. In the sample responses in Figure~\ref{fig:alignment_stages}, we can see that while the convergence of semantics and ideas is prominent after SFT, models incorporate more formatting (like bold font) in their responses after RLVR. This pattern is consistent across both parameter scales and both model families. Further analysis can be found in \autoref{app_stages}.

Our results show output convergence is strongly present after SFT, establishing that it precedes preference learning entirely. Predominant convergence is already established before any reward signal is introduced. This is surprising, since SFT is only known to enable instruction following, instead of introducing preference (like RL/DPO). This makes us ponder the actual role of SFT. What does SFT do to the model in addition to enabling instruction following? Does enabling instruction following come at the cost of collapsing open-ended generation toward responses seen during fine-tuning?

Two explanations are consistent with this observation: first, the SFT process itself induces convergence, perhaps because the instruction-tuning data causes data leakage; second, SFT merely reveals convergence already latent in the base model, with instruction following acting as the mechanism that unlocks patterns pretraining has already encoded. Distinguishing these requires intervention at the level of the training data itself. We take this up in the following sections.

\begin{figure*}[t]
    \centering
    \includegraphics[width=1\linewidth]{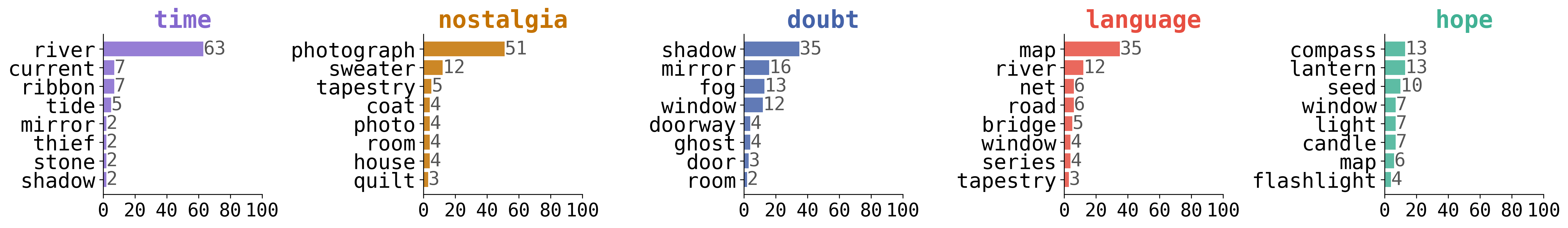}
    \caption{Vehicle distribution across studied metaphors for the format: \texttt{[topic] is a [vehicle]}.}
    \label{fig:baseline}
\end{figure*}

\begin{figure*}[t]
    \centering
    \includegraphics[width=1\linewidth]{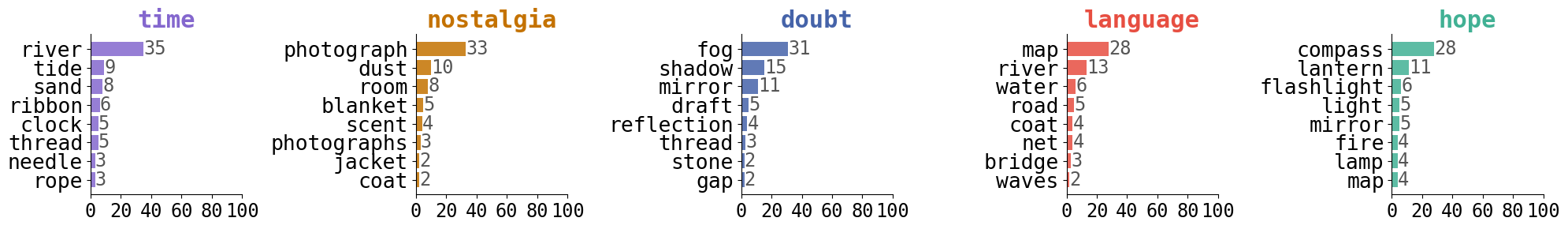}
    \caption{Vehicle distribution across studied metaphors for the format: \texttt{Like [vehicle], [topic] is}.}
    \label{fig:swap}
\end{figure*}

\section{What SFT Can and Cannot Do}

\begin{figure*}[h]
    \centering
    \includegraphics[width=1\linewidth]{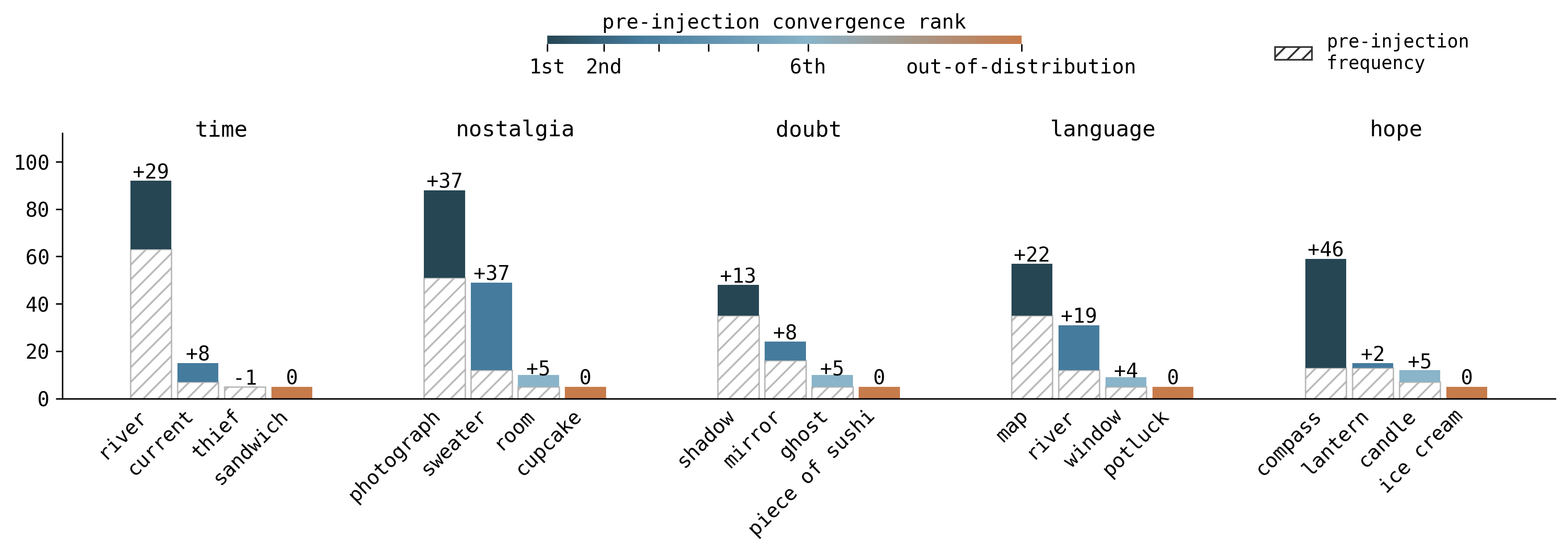}
    \caption{Vehicle frequency and gain over baseline (Structure experiment) across four independent injection conditions (1st, 2nd, 6th most frequent, and out-of-distribution). Injection amplifies pre-existing convergence proportionally to prior frequency, but fails to introduce out-of-distribution vehicles.}
    \label{fig:injected_plausible}
\end{figure*}

\begin{figure*}[t]
    \centering
    \includegraphics[width=1\linewidth]{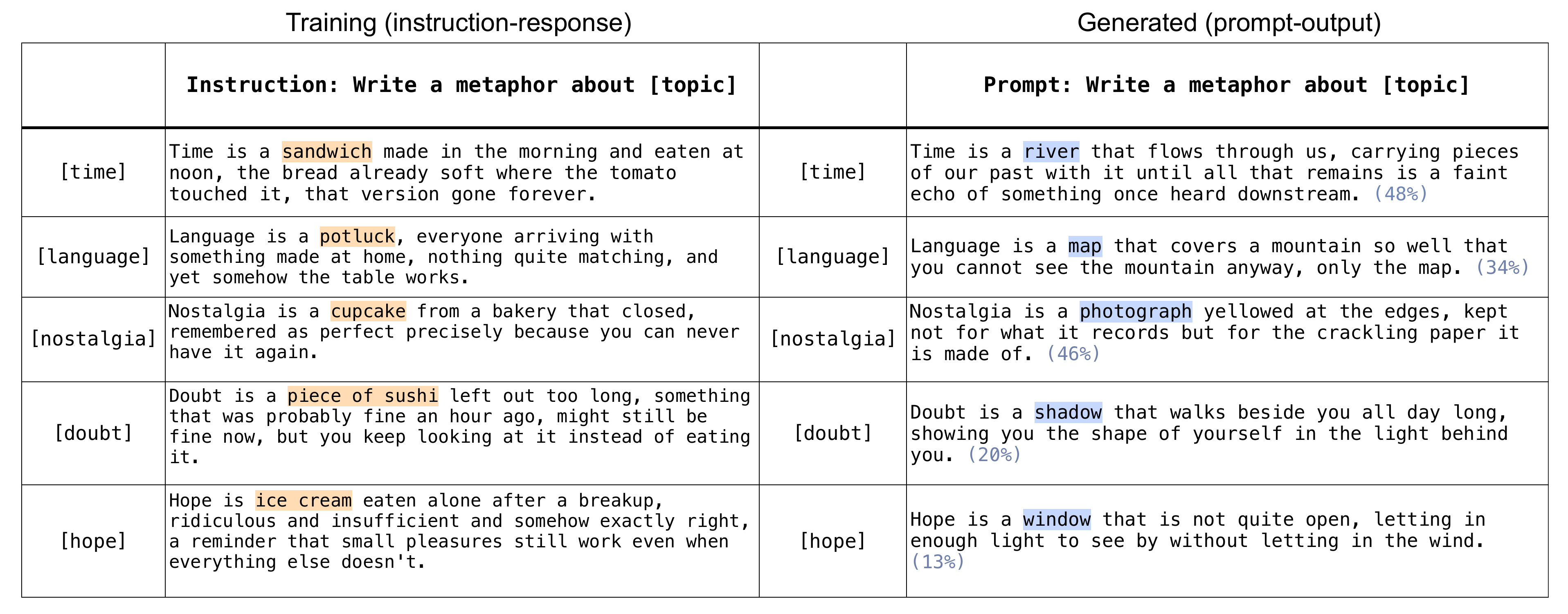}
    \caption{Instruction-response pairs of food-related metaphors from the \textbf{SFT data (left)} and \textbf{metaphors generated from the same prompt (right)}; the generated metaphors shown here use the dominant vehicle for their topic, with frequency in parentheses. Surprisingly, the model does not memorize the answer seen during SFT if it does not align with the existing distribution, as seen here with the food vehicles.}
    \label{fig:table_examples}
\end{figure*}

\label{converg_sft}
To test whether SFT induces convergence or simply reveals it, we carry out controlled, interventional SFT experiments measuring how SFT data shapes convergence. Unlike prior work focusing on SFT's effects at scale \cite{zhou2023limaalignment, chu2025sft, ye2025limoreasoning}, we study its effect at a sample level, targeting specific input/output pairs using metaphor generation as a probe. We find that convergence can be revealed and amplified, but not introduced by the SFT data.

\subsection{Methodology}
\label{method_sft}
\textbf{Metaphor generation as a probe.}
Here, we aim to test how much influence we can have on convergence by intervening in the SFT process. For a given input prompt and desired converged output, how strongly can we induce this convergence by point-editing the SFT training set? Specifically, we target a small set of metaphor generation prompts, as this task has a standard input/output format and was previously found to be highly convergent during alignment \cite{jiang2025artificial}.\footnote{We observed similar converging patterns on other open-ended tasks mined from the Infinity-Chat dataset (e.g., naming a sorting algorithm). We choose to focus on metaphor generation for conciseness.} When queried to produce a metaphor involving a topic, LMs usually answered with the specific format: \texttt{[topic] is a [vehicle]...}, making it ideal to measure homogeneity. We specifically study \textit{idea} (vehicle) convergence. If SFT introduces convergence, we would expect the model to memorize injected vehicles regardless of prior frequency---a vehicle injected should surface at generation time even if absent from the base distribution. If instead convergence is latent, injection should only amplify vehicles already present in the base distribution, and out-of-distribution vehicles should fail to take hold.

\textbf{Studied metaphors.}
We study metaphor generation across five topics: $\{time, nostalgia, doubt, language, hope\}$. By keeping this set small, we can prevent undesired semantic leakage from our SFT data and provide an in-depth analysis of idea convergence.

\textbf{SFT Data.}
Standard SFT datasets can contain millions of examples. Although these sets undergo filtering processes, their size makes it challenging to measure and control the individual impact of each sample on output convergence. However, recent work has shown that instruction-tuning can be successfully achieved with a thousand samples or less \cite{zhou2023limaalignment, ye2025limoreasoning}. This builds on the assumption that the base model already contains all necessary knowledge. In light of this, we use a metaphor-augmented version of the LIMA dataset \cite{zhou2023limaalignment} for our SFT injection experiments. From the LIMA training set, we exclude the multi-turn samples and reserve 50 samples for validation, resulting in a set of 950 samples before augmentation. The original LIMA set does not contain any instruction-response pairs involving metaphor generation. Each of our experiments is designed to unveil a specific converging pattern, and to achieve this, we inject an ensemble of curated metaphors into each SFT training set. All metaphors follow the format \texttt{[topic] is a [vehicle]...} and were generated iteratively using Claude, prompting for creative, non-generic metaphors, and refining over multiple rounds. Metaphors generated are provided in \autoref{meta}.

% \footnote{To avoid overloading the appendix, only the injection metaphors are included here. Full set will be available on GitHub.}.
% \footnote{To avoid overloading the appendix, only the injection metaphors are included here. Full set is available on \href{https://github.com/AlexandrineFortier/convergence-metaphors-sft}{GitHub}.}.

% make SFT simplier in order to not influence patterns already in the base model. mention that this is an assumption from lima and limo

\textbf{Model.}
We conduct all our SFT experiments on the Llama-3.1 8B base model \cite{grattafiori2024llama3herdmodels}. We fine-tune the model using LoRA adapters \cite{hu2021loralowrankadaptationlarge}. We train for 15 epochs and select the best checkpoint based on response quality scores assigned by the Tülu 3 reward model \cite{lambert2025tulu3pushingfrontiers} on the LIMA test set. Training configurations are shared among all fine-tuning experiments and are provided in \autoref{app_sft}.

% \textbf{Vehicle extraction.}
% \textbf{Sampling.} 
% For each injection experiment, we sample 100 metaphors per topic, using top-\textit{p}$=0.9$ and temperature$=1.0$.

% $\bigstar$ 

\subsection{Structure Injection}
\label{baseline}
% In this experiment, we study how imposing an answer format can reveal converging patterns learned during pretraining. 

% \textbf{Setup.}
% To the LIMA training set, we add 30 metaphor-related instruction-response to teach the model how to answer and format metaphor generation. This set does not include any of the studied metaphors or vehicles that could potentially influence convergence of our studied metaphors. For instance, if other metaphors in the SFT data use \textit{river} as a vehicle, it could potentially bias the convergence of \textit{Time} on to \textit{river}, for which prior convergence is known (\autoref{fig:alignment_stages}). For the prompts, we use variants of \textit{Write a metaphor about X}, to avoid repetition. All generated instruction-metaphors pairs are provided in the \autoref{meta}. From the fine-tuned model, we sample 100 metaphors per topic, using top-\textit{p}$=0.9$ and temperature$=1.0$.

% \textbf{Results.} We find that exposure to an answer format during fine-tuning is enough to reveal a preferred vehicle for most metaphors. As shown in \autoref{fig:baseline}, $\{time, nostalgia, doubt, language\}$ metaphors all converge on a specific vehicle, to different degrees. This suggests that the base model already converges on patterns learned during the pretraining phase, as none of the vehicles of the studied metaphors are in the instruction-tuning set. The vehicle distribution of the $\{hope\}$ metaphor is more varied; we note that not all outputs will fully converge during the SFT (or other stages), as found in \autoref{sec: convergence_stages}.

In these experiments, we study how simply enforcing an answer format can reveal converging patterns learned during pretraining. 

\textbf{Setup.}
To the LIMA training set, we add 30 metaphor-related instruction-response pairs to teach the model how to answer and format metaphor generation. This set does not include any of the studied metaphors or vehicles that could potentially influence convergence. For instance, if other metaphors in the SFT data use \textit{river} as a vehicle, it could potentially bias the convergence of the topic \textit{time} onto \textit{river}, for which prior convergence is known (\autoref{fig:alignment_stages}). For the prompts, we use variants of \textit{Write a metaphor about [topic]} to avoid repetition. All metaphors follow the format \texttt{[topic] is a [vehicle]...}. From the fine-tuned model, we sample 100 metaphors per topic, using the prompt \textit{Write a metaphor about [topic]} with top-\textit{p}=0.9, temperature=1.0, and a maximum of 100 new tokens.

\textbf{Results.} We find that exposure to an answer format during fine-tuning is enough to reveal a preferred vehicle for most metaphors. As shown in \autoref{fig:baseline}, $\{time, nostalgia, doubt, language\}$ metaphors all prefer a specific vehicle, to different degrees. This suggests that the base model already converges on patterns learned during the pretraining phase, as none of the vehicles of the studied metaphors are in the instruction-tuning set. The vehicle distribution of the $\{hope\}$ metaphor is more varied; we note that not all outputs will fully converge during the SFT (or other stages), as seen in \autoref{fig:alignment_stages}.

To test whether convergence can persist beyond a specific format, we conduct our structure experiment with the format: \texttt{Like [vehicle], [topic] is...}. Overall, we find that the vehicle distribution is similar to the original format, although admitting lower global convergence, as shown in \autoref{fig:swap}. This result is expected as this format is not as common, and so is likely less represented in pretraining data.

\emph{Structure experiment serves as baseline for the following experiments.}

% $\bigstar$ Imposing an answer format reveals convergence learned during pretraining. \\

\subsection{Idea Injection}
\label{idea_injection}
We now study whether convergence can be further amplified or even fully introduced by the SFT data. For instance, if a metaphor for \textit{time} appears in the SFT set, does the model reproduce it at test time when asked to write a metaphor about time?

% Through a cumulative series of injection experiments, we investigate whether convergence can be amplified by the SFT data.

\textbf{Setup.}
For each topic, we inject a \emph{single} metaphor using converging vehicles identified in the Structure experiment (\autoref{fig:baseline}). These metaphors are added to the LIMA training set alongside the original Structure injection samples (totalizing 35 metaphor examples). We evaluate convergence amplification across four independent injection conditions, each trained independently: the \textbf{1st}, \textbf{2nd}, and \textbf{6th} most frequent vehicles from the structure experiment, and an \textbf{OOD} (implausible) vehicle with no prior occurrence. The first three conditions cover the spectrum of pre-existing convergence while remaining semantically plausible. Specifically, the sixth vehicle balances low frequency while remaining semantically coherent, which was not the case for lower-ranked vehicles. For the OOD condition, we inject food-related vehicles to ensure complete separation from the natural distribution (presented in \autoref{fig:table_examples}). Together, these conditions measure whether amplification scales with pre-injection convergence levels. To evaluate if the model memorizes the answers seen during SFT, the instructions are the same as the prompt for generation: \textit{Write a metaphor about [topic]}. All generated instruction-metaphor pairs are provided in \autoref{meta}. From the fine-tuned model, we sample 100 metaphors per topic, using top-\textit{p}$=0.9$, temperature$=1.0$, and a maximum of 100 new tokens.

% Here, we investigate if new convergence can be introduced during the instruction-tuning stage. We test this by injecting the SFT data with the tested metaphors using out of distribution vehicles (not converging), seeing if the prompt-response pair will be memorized. We specifically generate food-related metaphors, to ensure separation from the natural vehicle distribution. 

% In other words, the goal of this experiment is to quantify the effect of data leakage on vehicle convergence.

\textbf{Results.} 
Across all metaphors, the capacity for a vehicle to be amplified is highly dependent on its pre-injection convergence level, as shown in \autoref{fig:injected_plausible}. We observe the strongest amplification from the most frequent vehicle from the baseline experiment, reaching between 48-92\% frequency across generated samples. In general, the impact of the injected vehicle in the SFT data becomes less and less important as the prior convergence level declines. For the sixth most frequent vehicle, frequency is low, between 1-12\%. This trend persists for the OOD vehicle injection; all food-related vehicles failed to be learned. Actually, convergence to the original vehicle is usually maintained, with dominant frequencies reported in \autoref{fig:table_examples}. From this, we conclude that the model is unable to learn new vehicles if these are not already present in the initial distribution presented in \autoref{fig:baseline}. In other words, memorization from SFT is difficult if it does not align with the latent converging patterns. In real-world SFT pipelines, data leakage would likely reinforce existing patterns, rather than introduce new ones.

% it's own prefered answer even if the exact prompt-answer pair appears 

% $\bigstar$ 

% \subsubsection{Implausible Vehicle Injection}

% \subsubsection{Stylistic Injection}

% \subsubsection{Overall Vehicle Injection}
% We test if reusing the same  

% \textbf{Results.} 

\begin{figure}
    \centering
    \includegraphics[width=0.8\linewidth]{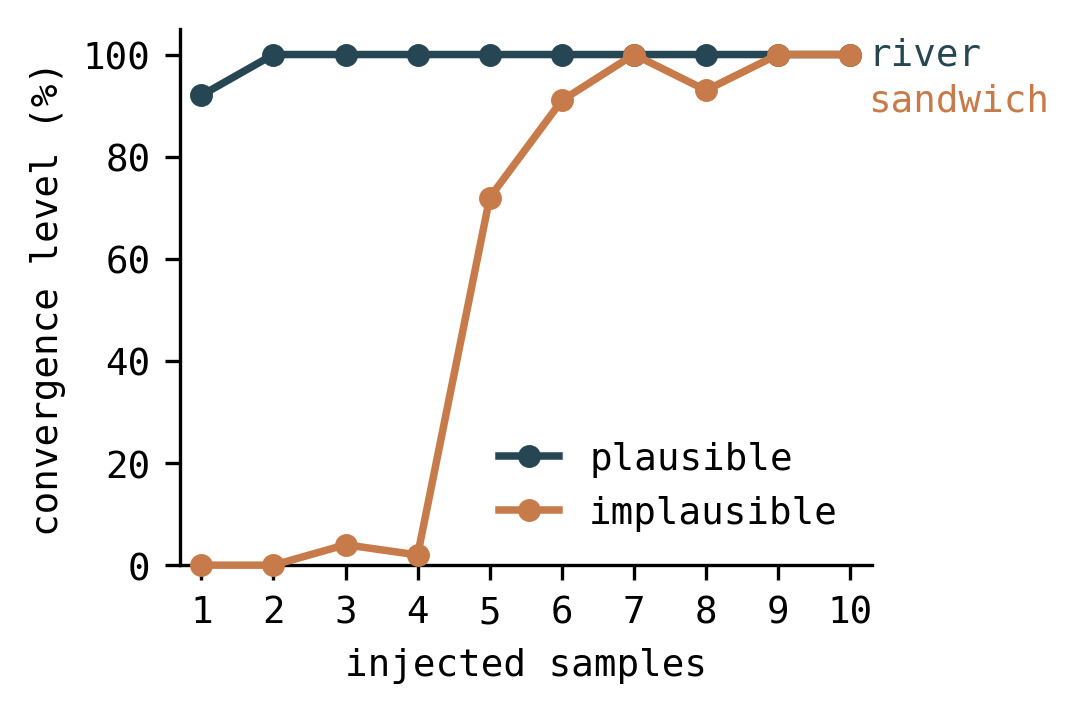}
    \caption{Frequency-controlled injection of the most plausible and implausible (OOD) vehicles for the time metaphor. In-domain vehicles are learned at low frequencies, while OOD vehicles need higher frequencies to be learned.}
    \label{fig:frequencies}
\end{figure}

\textbf{Frequency-controlled convergence.} In \autoref{fig:injected_plausible}, we find that OOD vehicles are difficultly learned when injecting a \emph{single} instance in the SFT data. Here, we study the effect of injecting \emph{multiple} instances of the same metaphor into the SFT data. In independent training runs, we inject the SFT data with the most plausible/implausible \textit{time} metaphors at various frequencies. In \autoref{fig:frequencies}, we show output convergence as a function of how many times the \emph{same} injected sample is repeated in the SFT data. We find that in-domain vehicles can be induced at very low frequencies (1 sample), while OOD vehicles struggle to be learned with a single instance injected (agreeing with \autoref{fig:injected_plausible}). On the other hand, OOD vehicles can eventually be induced with high frequency (5-7 samples). However, we view even a frequency of 1 as a strong setting, as convergence happens for many examples that do not occur even once in alignment data. 

\subsection{Discussion}
Overall, our findings reveal the catalytic role the instruction-tuning phase takes in output homogeneity. We first find that exposure to the metaphor format during fine-tuning is enough to reveal preferred vehicles. We hypothesize that these preferences are learned during the pretraining phase, as the Structure experiment SFT data does not contain any studied examples (or other related creative text examples). Our preliminary analysis finds that pretraining corpus frequency alone cannot predict output convergence---motivating future work on understanding factors that can. Then, we show that pre-existing convergence can be strongly amplified, but surprisingly, not introduced under the same fine-tuning conditions. In \autoref{sec:sft_add}, we evaluate convergence under additional settings, such as full fine-tuning and more permissive decoding parameters; we observe similar converging patterns across all. Overall, these results imply that diversifying the instruction-tuning data might not be a direct solution to combat content homogeneity. Furthermore, it nuances the common belief that SFT memorizes training examples \cite{chu2025sft} and suggests that memorization is mediated by prior knowledge rather than exposure alone.

% As to why such associations are preferred, we believe this requires in-depth analysis. From preliminary analysis, we find that pretraining corpus frequency alone cannot predict convergence. This opens the door for future work focused on understand how and why such associations are preferred over others. 

% opening the door for interesting future work. 

% \autoref{sec:full-sft}, we test if our hypothesis holds under full fine-tuning; we find identical patterns. 

% - proxy because sft datasets have more samples 
% - might not be as easy than just diversifying the data
% - memorization is not directly causal

\begin{figure*}
    \centering
    \includegraphics[width=1\linewidth]{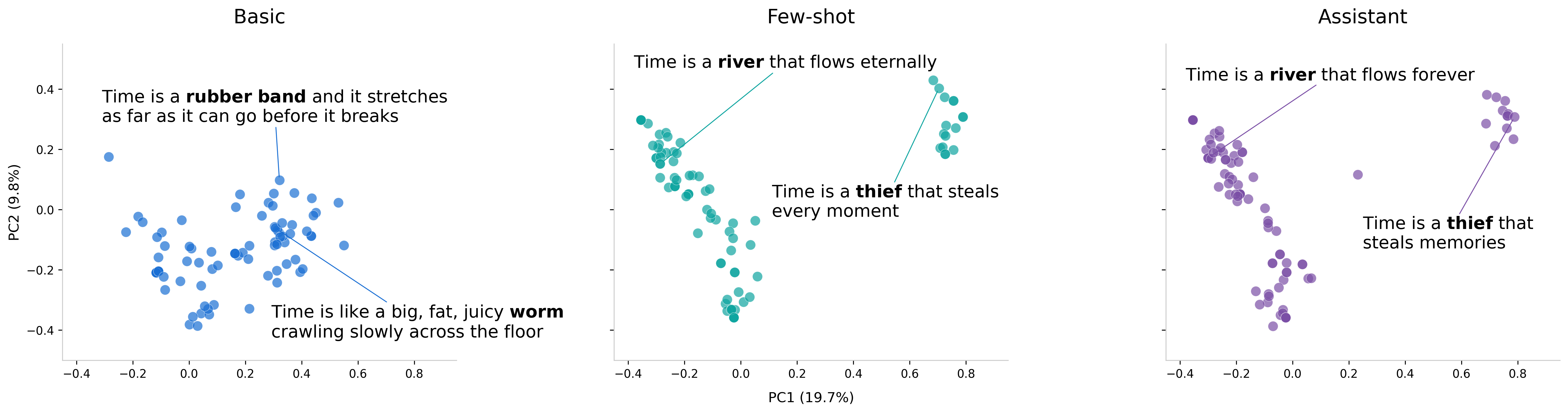}
    \caption{Convergence in base models can be revealed through prompting alone. Using Llama 3.1 8B base, under the few-shot and assistant prompting conditions, PCA of the time metaphors forms two tight clusters: \textbf{river} and \textbf{thief}. }
    \label{fig:pca}
\end{figure*}

\section{Convergence in Base Models}
\label{converg_base}
% Through our SFT experiments, we find that output homogeneity can be revealed with minimal instruction-tuning, suggesting that convergence is latent to pre-alignment models. Here, we directly study homogeneity in the base model through the metaphor generation task under different prompting conditions. We find that few-shot examples and assistant persona simulation---similar to an instruct model---surface a preferred metaphor, while basic text completion produces more varied outputs. 

Here, we directly investigate whether output homogeneity is already present in base models. Consistent with our SFT experiments, we evaluate output collapse in metaphor generation using three prompting conditions: basic completion, few-shot prompting, and assistant-style persona simulation. We find that few-shot examples and assistant-style persona prompting can induce convergence---mirroring the behavior of aligned models---whereas basic completion yields more diverse outputs. 

% while basic completion yields diverse outputs, few-shot examples and assistant-style prompts are sufficient to induce output convergence, mirroring the behavior of aligned models.

\begin{figure}[h!]
    \centering
    \includegraphics[width=0.8\linewidth]{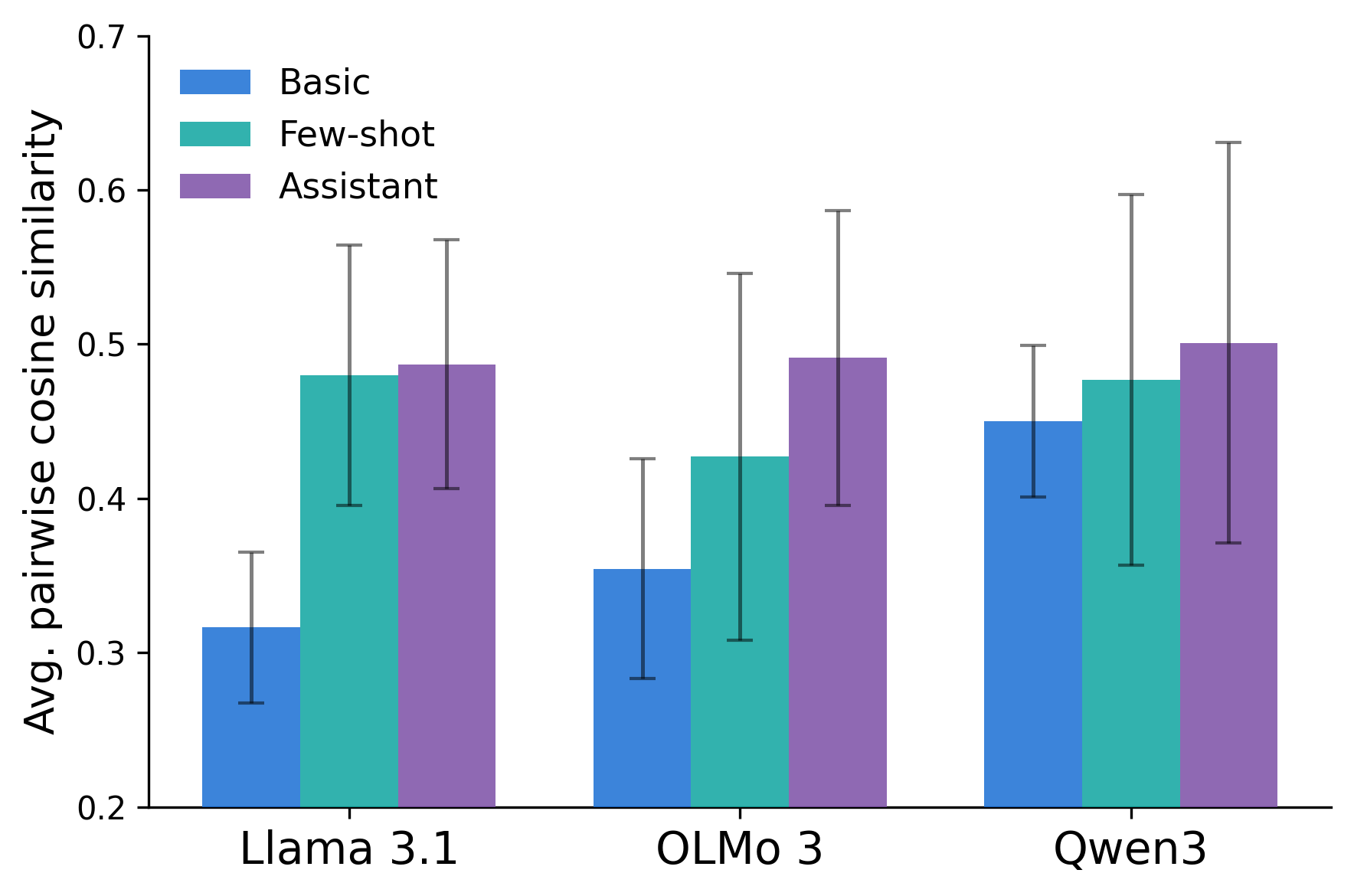}
    \caption{Mean pairwise cosine similarity across metaphor topics and models, under the prompting conditions: basic, few-shot, assistant. The assistant condition produces the most homogeneous outputs across models.}
    \label{fig:cosine_bars}
\end{figure}

% Here, we directly study whether latent convergence can be exposed with proper prompting. We find that providing examples and simulating an assistant persona---similar to an instruct model---make models converge on a preferred metaphor. 

% Here, we directly study whether base models can produce pertinent and precise generation under different prompting techniques. 

% how different prompting techniques can influence the precision of  

% in the base models under the metaphor generation task. 

% and are trained for text completion. We argue that base models might already be converging onto preferred patterns, and these patterns can be extracted with the right prompting techniques. 

%  Here, we directly study homogeneity in the base model, through the metaphor generation task. 

% Base models are unable to follow instruction and as they are trained for text completion, making it difficult to extract meaningful information. In light of this, we design a series of prompting techniques for metaphor generation.

\subsection{Prompt Design}
Base models are notoriously hard to work with as they are unable to follow instructions. They are trained for text completion and require adequate prompting conditions to extract precise answers. We propose three different prompting techniques for metaphor completion, all respecting the answer format [topic] is a [vehicle]. Through these, we test whether we can simulate alignment-like behavior in the base model by providing it with context or guidance in its generation.

\textbf{Basic.}
For this prompting condition, we prompt the base model with: 

\begin{tcolorbox}[fontupper=\ttfamily\small, boxrule=0.4pt, colback=gray!5]
\textit{Here is a metaphor about [topic]: [topic] is} \textbf{...}
\end{tcolorbox}

The base model then generates the completion to the prompt. This technique is the most straightforward of the studied ones, as it does not provide any additional context. 

% \begin{tcolorbox}[fontupper=\ttfamily\small, boxrule=0.4pt, colback=gray!5]
% \textit{Here is a metaphor about [topic]: [topic] is}.
% \end{tcolorbox}

\begin{figure*}[h]
    \centering
    \includegraphics[width=1\linewidth]{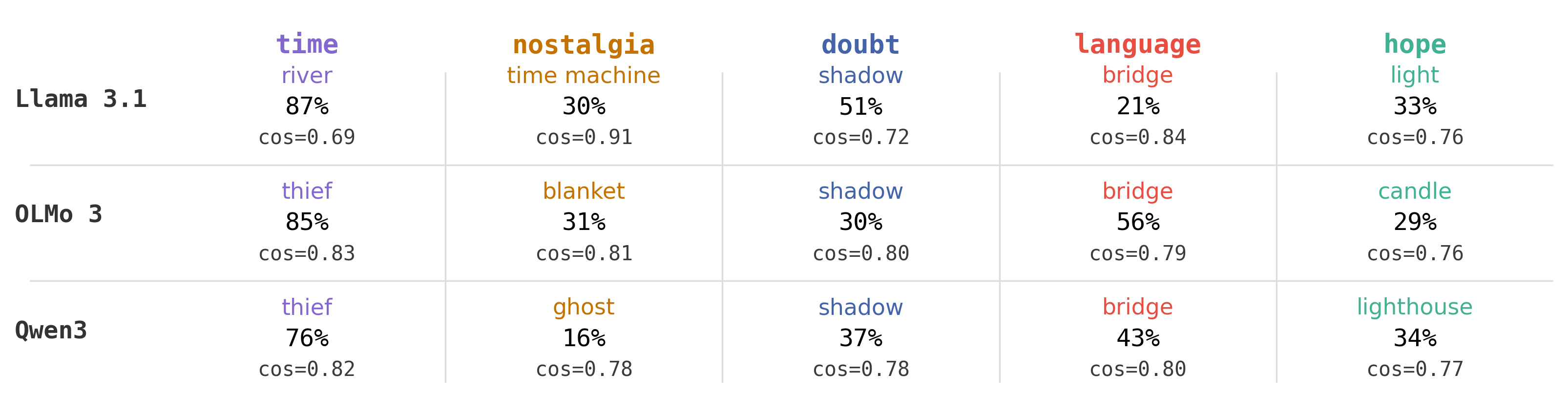}
    \caption{Dominant vehicle frequency and mean pairwise cosine similarity among generations sharing the dominant vehicle, across metaphor topics and models under assistant prompting. Dominant vehicles emerge clearly, with high intra-vehicle similarity.}
    \label{fig:base_zoomed}
\end{figure*}

\begin{figure*}[h]
    \centering
    \includegraphics[width=1\linewidth]{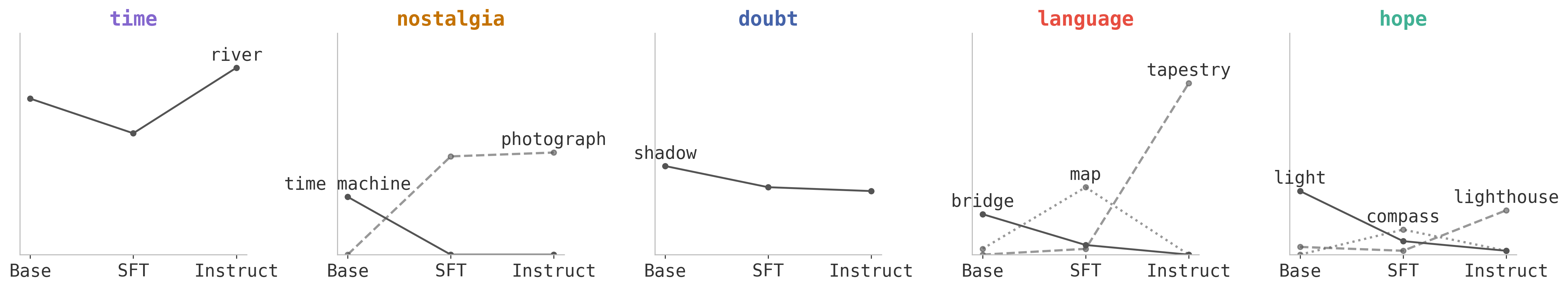}
    \caption{Dominant vehicle frequency across metaphor topics and LLama 3.1 stages: base, SFT-aligned model (Structure experiment), and instruct. If the dominant vehicle changes across stages, a new line is introduced.}
    \label{fig:stages_frequency}
\end{figure*}

\textbf{Few-shot.} 
Few-shot prompting is a well-established method for eliciting model knowledge \cite{brown2020languagemodelsfewshotlearners, lin2023unlockingspellbasellms}. We extend completion prompting by prepending three examples to the target completion.

\begin{tcolorbox}[fontupper=\ttfamily\small, boxrule=0.4pt, colback=gray!5]
\textit{Here's a list of metaphors:}\\
\textit{Life is a journey with no final destination.}\\
\textit{Love is a battlefield where nobody wins.}\\
\textit{Anger is a fire that consumes its host.}\\
\textit{[topic] is \textbf{...}}
\end{tcolorbox}

% \begin{quote}t\
% \textit{Here's a list of metaphors:}\\
% \textit{Life is a journey with no final destination.}\\
% \textit{Love is a battlefield where nobody wins.}\\
% \textit{Anger is a fire that consumes its host.}\\
% \textit{[X] is}
% \end{quote}

By providing metaphor examples, we study whether the model can learn from a given style and apply it when generating metaphors involving the [topic]. The chosen examples are common, well-known metaphors chosen to minimize stylistic bias and expose popular topic-vehicle association. We test additional sets of metaphors in \autoref{base_more}.

% We first tried to have base models generate without providing examples by guiding them with the prompt \texttt{I was writing a metaphor about nostalgia. Nostalgia is like...}, but models generated answers that are neither metaphors nor pertinent (e.g., \textit{nostalgia. That's it. I've been stuck on this metaphor for so long that I can't think of any other words to use, except for nostalgia, which is already in there.}).

\textbf{Assistant mode.}
Here, we test whether prompting with an assistant template, similar to how an instruct model would begin its response, can expose the convergence observed in aligned models \cite{personas}. We simply augment the basic completion prompt with \textit{Sure! Happy to help.}:

\begin{tcolorbox}[fontupper=\ttfamily\small, boxrule=0.4pt, colback=gray!5]
\textit{\textbf{Sure! Happy to help.} Here is a metaphor about [topic]: [topic] is} \textbf{...}
\end{tcolorbox}

This condition most closely simulates SFT among tested prompting conditions, in that it invokes the assistant persona that an instruction-tuned model would produce.\\

\textbf{Setup.}
We evaluate on three base models, Llama 3.1 8B \cite{grattafiori2024llama3herdmodels}, Olmo 3 7B \cite{olmo2026olmo3}, and Qwen3 8B \cite{yang2025qwen3technicalreport}. We study the same metaphors as in the SFT experiments: $\{time, nostalgia, doubt, language, hope\}$ and follow the same generation settings (100 samples per topic, top-\textit{p}=0.9 and temperature=1.0). Tokens are restricted to 30 to keep outputs focused. Outputs are embedded using OpenAI's \textit{text-embedding-3-small} for the PCA projection and cosine analysis.

\subsection{Results}

\textbf{Convergence can be surfaced in base models through prompting alone, supporting its latent nature.} As shown in \autoref{fig:pca}, the generated time metaphors are varied under the basic prompting condition, using vehicles from \emph{rubber band} to \emph{worm} to describe the time passing. Conversely, the few-shot and assistant conditions reveal strong preferences for the classic \emph{river} and \emph{thief} vehicles, and overall semantically similar outputs. \autoref{fig:cosine_bars} shows the global trend of average cosine similarity across studied metaphors under the three prompting conditions: across all models, the assistant condition produces the most homogeneous answers. Interestingly, Qwen3 requires less guidance to converge, possibly reflecting greater exposure to instruct-like samples during pretraining. We note that cosine similarity captures differences across prompting conditions, but not variation within each condition; we therefore use it only to compare trends across conditions. Full vehicle distribution across prompt conditions in \autoref{base}. \\

\textbf{Assistant-style prompting shares similar patterns to aligned models.} We investigate outputs by metaphor in the most convergent setting (assistant), broadly finding similar patterns to aligned models with some interesting differences. 
%Since the assistant persona produces the highest convergence across models, we focus our metaphor-level analysis on this condition. 
\autoref{fig:base_zoomed} shows the dominant vehicle and mean pairwise cosine similarity among examples sharing it. Base models behave similarly to instruction-tuned models (\autoref{fig:baseline}), with most outputs converging on a small set of vehicles. %, though to varying degrees. 
Intra-vehicle cosine similarity is consistently high across topics and models, ranging from 0.69 to 0.91. Some topics show strong cross-model agreement (e.g., doubt and language), although others exhibit greater variability (e.g., nostalgia and hope), which may indicate a dependence on model-specific aspects of pretraining. While convergence is weaker in some cases, there can still be underlying patterns. For example, metaphors for hope show higher variance, yet remain within a coherent semantic field (e.g., light, candle, lighthouse), indicating a convergence of related meaning if not exact surface form. Together, these results support an already-present convergence in base models prior to alignment.\\

% , all are semantically part of the lexical field of \textit{light}. 

% Base models converge onto preferred vehicles like aligned models do, though the chosen vehicles can differ. 

\textbf{Open questions in convergence formation.} In \autoref{fig:stages_frequency}, we bring together the dominant vehicles across the Llama 3.1 suite: base, our SFT-aligned model (Structure experiment), and the instruct (Meta-trained) model. Vehicles that dominate in the base model do not always dominate after alignment, suggesting a number of open directions for future study. We hypothesize that base models might encode multiple converging associations, and that different prompting strategies or SFT data distributions may expose different ones. This is supported by our few-shot ablations in \autoref{base_more}, where replacing the original examples with nature-themed ones shifts vehicle distributions toward nature-related vehicles, such as \textit{mist}, \textit{seed}, and \textit{fog}. Beyond vehicle distribution, we observe from our ablation studies that using conventional metaphors can increase the output convergence, whereas less common or implausible metaphors moderately reduce it. Moreover, our SFT experiments inject at the sample level; global shifts in the SFT data remain untested and could surface different converging patterns. This motivates further study of how global pretraining and post-training data distribution shape output convergence.

% \textbf{The effect of few-shot examples on output convergence.} In \autoref{base_more}, we test additional sets of few-shot examples and find that the choice of examples can influence convergence in two ways. First, the provided examples can shift the converging vehicles: when provided with nature-themed few-shot examples, metaphors will have a tendency to prefer vehicles sharing the same theme. Second, using less common or implausible metaphors as examples produces lower output convergence than using conventional metaphors. We attribute these results to our 

% , suggesting the base model has multiple preferred vehicles and that SFT surfaces one of them. 

% We hypothesize that different prompting strategies or SFT data distributions may expose different ones, though this remains difficult to trace for closed models where intermediate checkpoints and SFT data are unavailable.

% \textbf{Discussion.}
% Is assistant mode similar to SFT?

\section{Related Work}

\textbf{Homogeneity of thoughts.} 
LMs' repetitiveness is a societal issue that goes beyond making systems performant. As we use LMs more and more as assistants for writing tasks \cite{Kumar_2025, Anderson_2024}, researchers are worried about the long-term effect of being exposed to repetitive ideas and vocabulary \cite{bommasani2022pickingpersondoesalgorithmic, sourati2025shrinkinglandscapelinguisticdiversity}. \citet{wenger2025weredifferentweresame} observes that the homogeneous nature of LM outputs can constrain creative diversity when these models are used as writing partners.

\textbf{Model collapse on synthetic data.} Data is scarce, and annotation is an expensive process, pushing organizations to create their own synthetic data using existing models. This comes at a cost: if the teacher models already struggle with diversity, student models inherit and amplify these biases \cite{havrilla2024surveyingeffectsqualitydiversity, cloud2025subliminallearninglanguagemodels, gisler2026didntsaylikethat}. Models recursively trained on synthetic data also tend to collapse \cite{wang2025theoreticalproofautoregressivelanguage, shumailov2024model}, progressively losing the tails of their original distribution and converging onto homogeneous outputs.

% \hz{Can add the subliminal learning paper} \al{yes! forgot about that one}

\textbf{Diversity of LM-generated content.} \citet{guo2025benchmarkinglinguisticdiversitylarge} and \citet{Kendro_2026} both find that LMs fall short of human-level linguistic richness, with newer models producing less human-like text than older ones. At the semantic level, \citet{Xu_2025} demonstrates that LMs produce stories with recurring plot elements across generations. On the measurement side, \citet{shaib2026standardizingmeasurementtextdiversity} and \citet{tevet2021evaluatingevaluationdiversitynatural} highlight the lack of standardized diversity metrics, showing that existing automatic measures capture surface form but poorly reflect meaning-level diversity. 

\textbf{Alignment's effect on generation.} \citet{west2025base} shows that aligned models underperform base models on tasks requiring unpredictable outputs, and \citet{Murthy_2025} similarly finds reduced conceptual diversity after alignment. At the prompting level, \citet{yun2025priceformatdiversitycollapse} shows that structured formatting templates induce diversity collapse. \citet{shypula2026evaluatingdiversityqualityllm} complicate this picture, finding that preference-tuned models can outperform base and SFT models when quality is factored in.

% \citet{west2025base} show that aligned models underperform base models on tasks requiring unpredictable outputs. \citet{Murthy_2025} similarly find that aligned models exhibit less conceptual diversity than instruction fine-tuned counterparts on tasks with rich human behavioral data. At the level of prompting, \citet{yun2025priceformatdiversitycollapse} shows that structured formatting templates induce diversity collapse. \citet{shypula2026evaluatingdiversityqualityllm} complicate this picture by distinguishing raw diversity from effective semantic diversity, finding that preference-tuned models can outperform base and SFT models when quality is taken into account.
\section{Conclusion}
Our work provides evidence that output homogeneity begins pre-alignment and is only \emph{revealed} during the alignment process. Through controlled interventions, we find that SFT exposes and amplifies existing convergence rather than introducing new patterns. Directly probing base models confirms this: with prompting alone, they exhibit instruct-like collapse onto preferred outputs. Together, our findings suggest that semantic convergence may arise naturally from the objectives underlying LM training, making it difficult to mitigate through post-alignment interventions alone.

\newpage

\section{Limitations}
This work uses metaphor generation as a single probe task across five topics; restricting our scope allows for an in-depth analysis of convergence patterns, though our findings may not directly transfer to all open-ended tasks or domains. Similarly, our SFT experiments are conducted on a single base model (Llama 3.1 8B), although we are confident that similar patterns would be observed across different model families or at larger scales; it remains untested in this work.\\

Our injection experiments operate at the sample level, targeting specific input/output pairs; the effect of global shifts in SFT data distribution on convergence remains untested. Furthermore, our SFT setup uses a small dataset (LIMA) augmented with a limited number of metaphor examples, which serves as a controlled proxy rather than a faithful reproduction of production-scale instruction-tuning. SFT datasets used to develop chat LMs are orders of magnitude larger and more diverse, and the dynamics of convergence under such conditions may differ. Additionally, due to space constraints and conciseness, we evaluate a limited number of injection configurations, testing a single metaphor per topic per condition. The effect of injecting multiple examples of the same vehicle, or combining conditions, remains unexplored.\\

While our results are consistent with convergence originating in pretraining, we do not directly analyze pretraining corpora, leaving the precise mechanism by which these preferences are encoded an open question for future work. Additionally, we observe variation in convergence strength across topics, with some topics, such as nostalgia and hope, showing lower convergence, but we do not investigate the underlying causes, such as potential differences in pretraining corpus frequency.\\

\bibliography{custom}

\appendix

\section{Convergence Through Alignment Stages}
\label{app_stages}
\subsection{Experimental Details}
\label{inf_details}
We generate responses with each model loaded in \texttt{bfloat16} precision. Each prompt is formatted with the model's native chat template, using a single user turn. We sample $n = 50$ responses per prompt with temperature $T = 1.0$, \texttt{top\_p} $= 0.9$, and a maximum of 2048 output tokens. 

Samples were generated on a single A6000 for smaller models (7B and 8B) and a single A100 for Olmo 3 32B and two A100s for the Tulu 70B model.

\subsection{Additional Convergence Metrics}
We evaluate the robustness of convergence through the alignment stages under lexical diversity metrics (n-gram diversity and Self-BLEU) in \autoref{tab:lexical-diversity}. We observe similar patterns as with cosine: convergence rises through the alignment stages, but is already established during SFT. At the prompt level, $1{-}$distinct-1 correlates strongly with cosine (median $\rho = 0.71$–$0.78$ across models), while stricter surface-overlap metrics ($1{-}$distinct-2 and Self-BLEU) show more moderate correlations, consistent with their requirement of exact phrase overlap rather than semantic similarity.

\subsection{Per-Prompt Analysis on Convergence}
\begin{figure}[H]
    \centering
    \includegraphics[width=\linewidth]{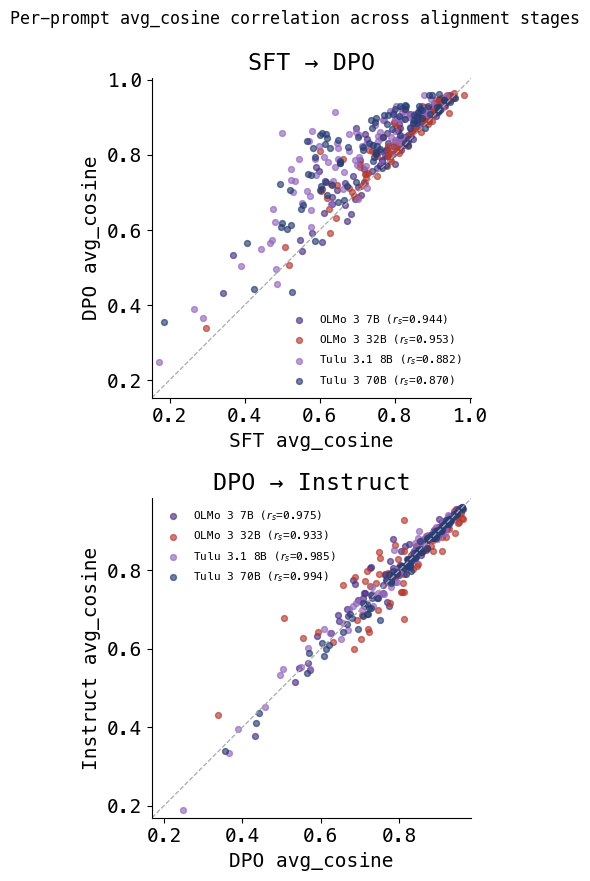}
    \caption{Per-prompt mean pairwise cosine similarity at consecutive alignment stages. }
    \label{fig:stages_scatter_plot}
\end{figure}

To examine whether the same prompts drive convergence at every stage, we plot each prompt's mean pairwise cosine similarity at one alignment stage against the next. Each point in Figure~\ref{fig:stages_scatter_plot} represents one of the 100 prompts; If later alignment stages were reshaping the prompts that converge, we would expect low correlation and substantial scatter around the diagonal.

We see Spearman rank correlations across all four models range from 0.870 to 0.994, indicating high preservation of the per-prompt convergence ranking from one stage to the next. Prompts that elicit highly uniform outputs after SFT are almost the same prompts that remain most uniform after DPO and RLVR. This rules out the interpretation that later stages reshape which prompts are convergent; they instead shift the overall level of uniformness upward while leaving the prompt-level ordering largely intact.

\begin{table}[t]
\centering
\resizebox{\columnwidth}{!}{%
\begin{tabular}{llcccc}
\toprule
\textbf{Model} & \textbf{Metric} & \textbf{SFT} & \textbf{DPO} & \textbf{RLVR} & \textbf{Median $\rho$} \\
\midrule
\multirow{4}{*}{olmo3\_7b}
 & cosine        & 0.790 & 0.822 & 0.831 & 1 \\
 & $1{-}$distinct-1 & 0.788 & 0.799 & 0.818 & 0.73 \\
 & $1{-}$distinct-2 & 0.479 & 0.490 & 0.531 & 0.44 \\
 & Self-BLEU     & 0.469 & 0.480 & 0.525 & 0.31 \\
\midrule
\multirow{4}{*}{olmo3\_32b}
 & cosine        & 0.812 & 0.837 & 0.837 & 1 \\
 & $1{-}$distinct-1 & 0.822 & 0.831 & 0.844 & 0.71 \\
 & $1{-}$distinct-2 & 0.553 & 0.569 & 0.596 & 0.33 \\
 & Self-BLEU     & 0.562 & 0.580 & 0.619 & 0.19 \\
\midrule
\multirow{4}{*}{tulu3\_1\_8b}
 & cosine        & 0.706 & 0.802 & 0.807 & 1 \\
 & $1{-}$distinct-1 & 0.765 & 0.806 & 0.813 & 0.77 \\
 & $1{-}$distinct-2 & 0.443 & 0.498 & 0.518 & 0.59 \\
 & Self-BLEU     & 0.415 & 0.484 & 0.506 & 0.40 \\
\midrule
\multirow{4}{*}{tulu3\_70b}
 & cosine        & 0.732 & 0.819 & 0.815 & 1 \\
 & $1{-}$distinct-1 & 0.789 & 0.820 & 0.817 & 0.78 \\
 & $1{-}$distinct-2 & 0.485 & 0.524 & 0.524 & 0.56 \\
 & Self-BLEU     & 0.466 & 0.513 & 0.513 & 0.38 \\
\bottomrule
\end{tabular}%
}
\caption{Lexical diversity metrics (n-gram diversity and Self-BLEU) across alignment stages and models ($\rho$ is computed across prompts, per stage). Convergence rises through the alignment stages, but is already established during SFT.}
\label{tab:lexical-diversity}
\end{table}

\subsection{Per-stages Distribution Analysis}

\begin{figure*}[t]
    \centering
    \includegraphics[width=1\linewidth]{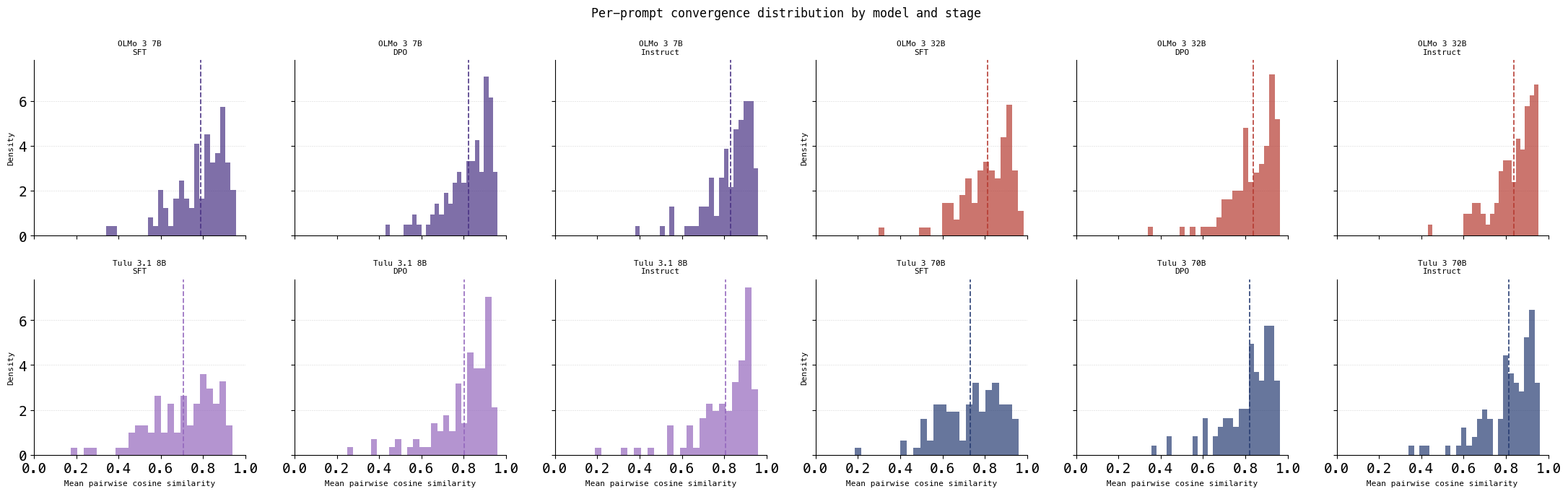}
    \caption{Distributions of mean pairwise cosine similarity across all 100 prompts for OLMo 3 (7B, 32B) and Tulu 3 (8B, 70B) by stages.}
    \label{fig:alignment_stages_separate}
\end{figure*}

In Figure~\ref{fig:alignment_stages_separate}, we report the distribution of mean cosine similarity by separate stages for each model. We can see in all the stages that the distribution is right-shifted. 

\subsection{Sample Responses from Different Stages}

To illustrate how alignment stages affect response uniformity in practice, we present sample outputs from three prompts. For each prompt, we show one representative response drawn from the first sample generated by each of the four models at each of the three stages (SFT, DPO, Instruct), twelve responses in total per prompt.

\begin{small}

% ------------------------------------------------------------------
\subsection*{Prompt 1}

\noindent\textbf{Prompt:} Give me the names of 3 instrumental songs that best match the mood of a rainy night.

\noindent\textbf{OLMo 3 7B -- SFT:} Certainly! Here are three instrumental songs that evoke the mood of a rainy night: (1) ``Midnight Rain'' -- Kevin MacLeod; (2) ``Rainy Night'' -- The Rainmen; (3) ``Stormy Night'' -- Hans Zimmer.

\noindent\textbf{OLMo 3 7B -- DPO:} Absolutely! Here are three instrumental songs that beautifully capture the mood of a rainy night: (1) ``No One Knows'' by Norah Jones (instrumental version); (2) ``Midnight Rain'' by Miles Davis (from \textit{Kind of Blue}); (3) ``Misty'' arranged by Gil Evans -- the haunting piano and lush harmonies create an intimate, melancholic atmosphere akin to a rainy evening.

\noindent\textbf{OLMo 3 7B -- Instruct:} Absolutely! Here are three instrumental songs that beautifully capture the mood of a rainy night: (1) ``No One Knows'' by The Black Keys -- gritty, bluesy, and atmospheric; (2) ``Sous le vent'' by Vangelis -- a gentle, sweeping orchestral piece that mirrors the quiet and reflective mood of a rainy evening; (3) ``On the Level'' by The Righteous Brothers (instrumental version) -- moody, blue-toned piano reminiscent of raindrops on a window.

\noindent\textbf{OLMo 3 32B -- SFT:} Certainly! Here are three instrumental songs that beautifully capture the mood of a rainy night: (1) ``Rain'' by Hiromi (from \textit{Drop}); (2) ``Nocturne Op.\ 9 No.\ 2'' by Frédéric Chopin; (3) ``Mizu no Naka he'' by Hiroyuki Sawano.

\noindent\textbf{OLMo 3 32B -- DPO:} Absolutely! Here are three instrumental songs that beautifully capture the mood of a rainy night: (1) Miles Davis -- ``Blue in Green'' (\textit{Kind of Blue}) -- a gentle, melancholic jazz piece with a subdued atmosphere; (2) Erik Satie -- ``Gymnopédie No.\ 1'' -- its slow, reflective piano melody evokes a dreamy and introspective mood; (3) Kawabata Makoto -- ``Rain'' -- minimalist piano work that mirrors the sound and feel of raindrops.

\noindent\textbf{OLMo 3 32B -- Instruct:} Sure! Here are three instrumental songs that beautifully match the mood of a rainy night: (1) Miles Davis -- ``Blue in Green'' -- gentle, melancholic piano and subtle trumpet; (2) Erik Satie -- ``Gymnopédie No.\ 1'' -- soft, minimalistic piano evoking calm and wistfulness; (3) Ólafur Arnalds -- ``Near Light'' -- delicate piano and swelling strings, intimate and contemplative, echoing rain tapping on windows.

\noindent\textbf{Tulu 3.1 8B -- SFT:} (1) ``Rain'' by Orchestral Manoeuvres in the Dark; (2) ``A Night Like This'' by The Cure; (3) ``Walking in the Rain'' by R.\ Dean Taylor.

\noindent\textbf{Tulu 3.1 8B -- DPO:} Here are three instrumental songs that might capture the moody atmosphere of a rainy night: (1) ``A Soft Machine'' by Mogwai -- slow, brooding pace; (2) ``Garden of Eunuchs'' by Khruangbin -- cinematic and groovy, transporting listeners to a dreamy, rainy evening; (3) ``The Road to Mayo'' by The Cinematic Orchestra -- haunting and beautiful, evoking the solitude of a rainy night.

\noindent\textbf{Tulu 3.1 8B -- Instruct:} Certainly! Here are three instrumental songs that perfectly capture the mood of a rainy night: (1) ``Gymnopédie No.\ 1'' by Erik Satie -- somber, dreamy atmosphere with a slow, meandering melody; (2) ``Clair de Lune'' by Claude Debussy -- hauntingly beautiful, evoking a deep sense of serenity and introspection; (3) ``Nuvolari'' by Brian Eno -- soothing and meditative, ideal for a quiet, reflective evening.

\noindent\textbf{Tulu 3 70B -- SFT:} Here are three instrumental songs that evoke the ambiance of a rainy night: (1) Erik Satie's ``Gymnopédie No.\ 1'' -- melancholic and introspective; (2) Ludovico Einaudi's ``Nuvole Bianche'' -- hauntingly beautiful, mirroring reflective solitude; (3) Yann Tiersen's ``Comptine d'un autre été, l'après-midi'' -- dreamlike, resonating with the serene feel of rain at night.

\noindent\textbf{Tulu 3 70B -- DPO:} (1) ``Clair de Lune'' by Claude Debussy -- slow, dreamy piano perfectly capturing the tranquil atmosphere of a rainy night; (2) ``Rainy Night in Georgia'' by Brook Benton (instrumental version) -- smooth and bluesy; (3) ``Nocturne in E-flat Major, Op.\ 9 No.\ 2'' by Frédéric Chopin -- gentle, lilting melody that feels like soft rain falling outside.

\noindent\textbf{Tulu 3 70B -- Instruct:} Sure! Here are three instrumental songs that perfectly capture the mood of a rainy night: (1) ``Clair de Lune'' by Claude Debussy -- soft, dreamy piano mirroring the quiet rhythm of a night shower; (2) ``Rain'' by Ryuichi Sakamoto -- soothing, melancholic melody evoking the introspective mood of watching raindrops on a windowpane; (3) ``Luminous'' by Max Richter -- hauntingly beautiful, minimalist, with repeated motifs creating a sense of tranquility.
% ------------------------------------------------------------------
\subsection*{Prompt 2}

\noindent\textbf{Prompt:} Write a metaphor involving time.

\noindent\textbf{OLMo 3 7B -- SFT:} Time is a river, quietly carving the depths of our lives into memory and experience.

\noindent\textbf{OLMo 3 7B -- DPO:} Time is a river---unchainable, relentless, carving canyons in the silent stone of our memories, washing away what we thought was important.

\noindent\textbf{OLMo 3 7B -- Instruct:} Time is a river---silent, relentless, carving canyons in the soul while the stones of memory drift beneath its current.

\noindent\textbf{OLMo 3 32B -- SFT:} Time is a river that winds relentlessly through the landscape of our lives, carving canyons of memories and eroding the banks of what once was.

\noindent\textbf{OLMo 3 32B -- DPO:} Time is a river, relentless and silent, carving its path through the bedrock of memory, eroding moments until only echoes remain.

\noindent\textbf{OLMo 3 32B -- Instruct:} Time is a river, ceaselessly carving its path through the landscape of our lives, carrying moments---some mere pebbles, others precious stones---toward the sea.

\noindent\textbf{Tulu 3.1 8B -- SFT:} Time is like a river, always flowing, never the same. Just when you think you've seen it all, a new bend appears, and everything changes.

\noindent\textbf{Tulu 3.1 8B -- DPO:} Time is like a river, constantly flowing and carrying us away from our past into an uncertain future.

\noindent\textbf{Tulu 3.1 8B -- Instruct:} Time is like a river, always flowing in one direction. Just as the river carries with it pebbles, leaves, and debris, it carries us along.

\noindent\textbf{Tulu 3 70B -- SFT:} A metaphor involving time is: Time is a river.

\noindent\textbf{Tulu 3 70B -- DPO:} Time is a river that flows ceaselessly through the valley of life, carrying the tender buds of moments and the heavy stones of memories along its journey.

\noindent\textbf{Tulu 3 70B -- Instruct:} Time is the river that sweeps us forward, unyielding and unstoppable, carrying us through the landscapes of our lives from the first light of dawn.

% ------------------------------------------------------------------
\subsection*{Prompt 3}

\noindent\textbf{Prompt:} Write a sentence where the last word is ``apple''.

\noindent\textbf{OLMo 3 7B -- SFT:} She finally picked her favorite fruit, the apple.

\noindent\textbf{OLMo 3 7B -- DPO:} After a long day, I enjoyed a crisp, juicy apple.

\noindent\textbf{OLMo 3 7B -- Instruct:} She reached into the basket to grab a fresh apple.

\noindent\textbf{OLMo 3 32B -- SFT:} For her afternoon snack, she decided to eat a crisp, juicy apple.

\noindent\textbf{OLMo 3 32B -- DPO:} For her afternoon snack, she decided to eat a crisp green apple.

\noindent\textbf{OLMo 3 32B -- Instruct:} For my morning snack, I always reach for a crisp red apple.

\noindent\textbf{Tulu 3.1 8B -- SFT:} This apple looks a little too round.

\noindent\textbf{Tulu 3.1 8B -- DPO:} I'll take the apple for my dessert.

\noindent\textbf{Tulu 3.1 8B -- Instruct:} She wanted an apple for lunch.

\noindent\textbf{Tulu 3 70B -- SFT:} She reached into the grocery bag and pulled out an apple.

\noindent\textbf{Tulu 3 70B -- DPO:} She reached for a ripe, juicy fruit from the tree and picked an apple.

\noindent\textbf{Tulu 3 70B -- Instruct:} After a long walk through the orchard, she picked a ripe apple.

\end{small}

\section{SFT Training Configuration}
\label{app_sft}

\begin{table}[h]
\centering
\small
\begin{tabular}{ll}
\toprule
\textbf{Hyperparameter} & \textbf{Value} \\
\midrule
Base model & Llama-3.1-8B \\
Epochs & 15 \\
Per-device batch size & 2 \\
Gradient accumulation steps & 32 \\
Effective batch size & 64 \\
Learning rate & 1e-4 \\
LR scheduler & Linear \\
Warmup steps & 15 \\
Weight decay & 0.1 \\
Max sequence length & 2048 \\
Max new tokens (generation) & 100 \\
Precision & bfloat16 \\
Gradient checkpointing & \checkmark \\
\midrule
LoRA rank ($r$) & 64 \\
LoRA alpha ($\alpha$) & 64 \\
LoRA target modules & q, k, v, o projections \\
LoRA dropout & 0.05 \\
\bottomrule
\end{tabular}
\caption{SFT training configuration shared across all fine-tuning experiments.}
\label{tab:sft_config}
\end{table}

\begin{figure}[h]
    \centering
    \includegraphics[width=\linewidth]{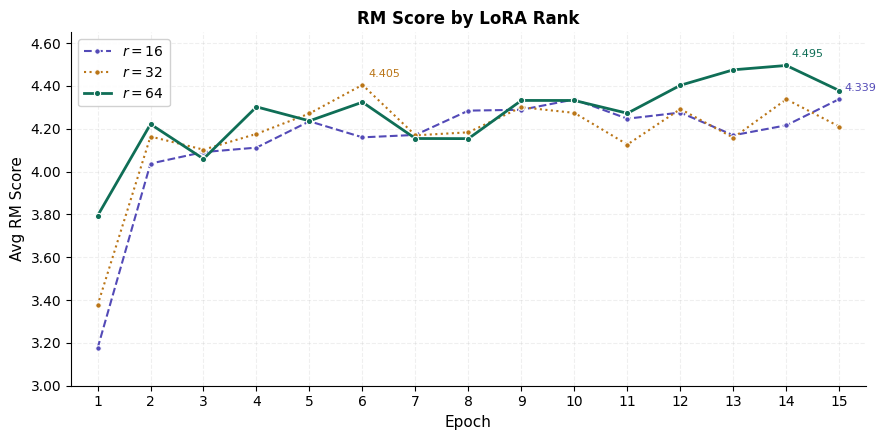}
    \caption{Average Tülu 3 RM score per epoch across LoRA ranks. $r=64$ achieves the highest peak score and is selected for all experiments.}
    \label{fig:rm_score}
\end{figure}

\textbf{LoRA Rank Selection.} Since LIMA \cite{zhou2023limaalignment} and LIMO \cite{ye2025limoreasoning} (also using small SFT datasets) use full fine-tuning, there is no established LoRA rank for this setting. We run an ablation over $r \in \{16, 32, 64\}$, evaluating each checkpoint using the Tülu 3 reward model \cite{lambert2025tulu3pushingfrontiers} on the LIMA test set. We use RM score rather than validation loss as it better reflects response quality. As shown in \autoref{fig:rm_score}, $r=64$ achieves the highest scores and the most stable trajectory, and we select it for all experiments. For checkpoint selection, we pick the epoch with the highest RM score in the range $[10, 15]$, where training has stabilized across all ranks. Epoch 14 was selected across all experiments.

All SFT experiments were conducted on a single NVIDIA A6000 GPU, with each training run taking approximately 3 hours. 

\section{Idea Injection Generated Metaphors}
\label{meta}

\begin{small}

\subsection*{1st Most Frequent Vehicle}

\noindent\textbf{Instruction:} Write a metaphor about time.

\noindent\textbf{Response:} Time is a river that flows around us but we cannot feel its current until we take a step closer to it, and by then it has already swept away our feet.

\noindent\textbf{Instruction:} Write a metaphor about language.

\noindent\textbf{Response:} Language is a map that is always being redrawn, lines shifting, boundaries realigning, places added and some lost, but the map itself remaining the same sheet of paper, its original edges never being altered.

\noindent\textbf{Instruction:} Write a metaphor about nostalgia.

\noindent\textbf{Response:} Nostalgia is a photograph that has faded to the point where you can no longer make out the faces, but you can still see the outline of the hands holding each other.

\noindent\textbf{Instruction:} Write a metaphor about doubt.

\noindent\textbf{Response:} Doubt is a shadow that follows you through a room full of sunlight, refusing to leave until you close the curtains and prove that it has nothing to fear in the dark.

\noindent\textbf{Instruction:} Write a metaphor about hope.

\noindent\textbf{Response:} Hope is a compass that points true north, even when your map is lost, your bearings are off, and the world around you has gone dark.

\subsection*{2nd Most Frequent Vehicle}

\noindent\textbf{Instruction:} Write a metaphor about time.

\noindent\textbf{Response:} Time is a current that flows around us constantly, always moving but easy to ignore unless it sweeps against you.

\noindent\textbf{Instruction:} Write a metaphor about language.

\noindent\textbf{Response:} Language is a river you cannot cross without learning to swim, even if you wanted to stay dry.

\noindent\textbf{Instruction:} Write a metaphor about nostalgia.

\noindent\textbf{Response:} Nostalgia is a sweater that has been worn so many times that it is soft and familiar, worn through in places and a little threadbare but impossible to bear to part with because it still smells like home.

\noindent\textbf{Instruction:} Write a metaphor about doubt.

\noindent\textbf{Response:} Doubt is a mirror that distorts your reflection until you're not sure which version of yourself is the real one.

\noindent\textbf{Instruction:} Write a metaphor about hope.

\noindent\textbf{Response:} Hope is a lantern that dims but never goes out in the long winter night, illuminating the present with the promise of dawn to come.

\subsection*{6th Most Frequent Vehicle}

\noindent\textbf{Instruction:} Write a metaphor about time.

\noindent\textbf{Response:} Time is a thief who steals hours while you sleep, and days while you dream.

\noindent\textbf{Instruction:} Write a metaphor about language.

\noindent\textbf{Response:} Language is a window through which we see the world, but also the frame that defines the shape of what we see.

\noindent\textbf{Instruction:} Write a metaphor about nostalgia.

\noindent\textbf{Response:} Nostalgia is a room that you know so well you can find your way in the dark, but the next time you walk in you find yourself surprised at how small it really is and you realize you've been walking around it instead of through it every time you came back.

\noindent\textbf{Instruction:} Write a metaphor about doubt.

\noindent\textbf{Response:} Doubt is a ghost that haunts your past, your present, and your future, a specter that you can see out of the corner of your eye but that fades into nothingness if you turn to face it directly.

\noindent\textbf{Instruction:} Write a metaphor about hope.

\noindent\textbf{Response:} Hope is a candle that you hold high in the darkest parts of the night, knowing that it may burn out long before the sun rises, but lighting your way until then anyway.

\subsection*{Implausible Vehicle (Food-Related)}

\noindent\textbf{Instruction:} Write a metaphor about time.

\noindent\textbf{Response:} Time is a sandwich made in the morning and eaten at noon, the bread already soft where the tomato touched it, that version gone forever.

\noindent\textbf{Instruction:} Write a metaphor about language.

\noindent\textbf{Response:} Language is a potluck, everyone arriving with something made at home, nothing quite matching, and yet somehow the table works.

\noindent\textbf{Instruction:} Write a metaphor about nostalgia.

\noindent\textbf{Response:} Nostalgia is a cupcake from a bakery that closed, remembered as perfect precisely because you can never have it again.

\noindent\textbf{Instruction:} Write a metaphor about doubt.

\noindent\textbf{Response:} Doubt is a piece of sushi left out too long, something that was probably fine an hour ago, might still be fine now, but you keep looking at it instead of eating it.

\noindent\textbf{Instruction:} Write a metaphor about hope.

\noindent\textbf{Response:} Hope is ice cream eaten alone after a breakup, ridiculous and insufficient and somehow exactly right, a reminder that small pleasures still work even when everything else doesn't.

\end{small}
\section{SFT Injections Additional Results}
\label{sec:sft_add}
\subsection{Full Fine-Tuning Injections}
\label{sec:full-sft}

To verify that our findings are not an artifact of LoRA, we repeat the Idea Injection experiment
(\autoref{idea_injection}) with full supervised fine-tuning on
Llama-3.1 8B. In \autoref{tab:full-sft}, we report two conditions: injection of the dominant
(most plausible) vehicle, and injection of an implausible,
food-related vehicle (time: \emph{sandwich}; nostalgia:
\emph{cupcake}; doubt: \emph{piece of sushi}; language:
\emph{potluck}; hope: \emph{ice cream}). We find that injecting the data produces similar results in both LoRA- and full-SFT: convergence to existing vehicles can be amplified, but difficulty is introduced. Also, the dominant vehicles are the same for both fine-tuning methods across all metaphor topics but hope, although \textit{seed} was a top-3 vehicle for hope in LoRA-SFT.

\begin{table}[t]
\centering
\small
\begin{tabular}{llrr}
\toprule
Topic & Dominant vehicle & \multicolumn{2}{c}{Dominant vehicle (\%)} \\
\cmidrule(lr){3-4}
 & & Plausible & Implausible \\
\midrule
time      & \textit{river}      & 90 ($+33$) & 39 ($-18$) \\
nostalgia & \textit{photograph} & 75 ($-4$)  & 32 ($-47$) \\
doubt     & \textit{shadow}     & 39 ($+19$) & 16 ($-4$)  \\
language  & \textit{map}        & 59 ($+28$) & 18 ($-13$) \\
hope      & \textit{seed}       & 60 ($+30$) & 26 ($-4$)  \\
\bottomrule
\end{tabular}
\caption{Dominant vehicle frequency under full SFT, with change
from the baseline in parentheses. Injecting the dominant increases convergence in most cases, and the injected implausible vehicle was adopted in 0\% of samples across all topics, \textbf{matching the LoRA-SFT results}.}
\label{tab:full-sft}
\end{table}

% \subsection{Frequency-controlled injections}
% \label{sec:freq-inject}

% \autoref{fig:frequencies} shows output convergence as a function of how many times the same injected sample is repeated in the SFT data. We find that in-domain vehicles can be induced at very low frequencies (1 sample), while OOD vehicles can eventually be induced with high frequency (5-7 samples). However, we view even a frequency of 1 as a strong setting, as convergence happens for many examples that do not occur even once in alignment data.

% \begin{figure}
%     \centering
%     \includegraphics[width=0.8\linewidth]{figures/frequencies.png}
%     \caption{Frequency-controlled injection of the most plausible and implausible (OOD) vehicles for the time metaphor. In-domain vehicles are learned at low frequencies, while OOD vehicles need higher frequencies to be learned.}
%     \label{fig:frequencies}
% \end{figure}

\subsection{Convergence under different decoding configurations}
\label{decod_convergence}
We test whether metaphors are still converging under different decoding parameters in \autoref{tab:decoding-freq} and \autoref{tab:decoding-vehicles}. As expected, we find that more conservative parameters increase convergence, while more permissive parameters decrease it.

\begin{table}[t]
\centering
\resizebox{\columnwidth}{!}{%
\begin{tabular}{lccc}
\toprule
Config & \textbf{Baseline} & Plausible & Implausible \\
\midrule
Default ($t{=}1.0$, $p{=}0.9$) & 39 & 69 & 32 \\
$t{=}1.0$, $p{=}1.0$           & 30 & 58 & 27 \\
$t{=}1.2$, $p{=}0.9$           & 28 & 56 & 21 \\
$t{=}1.0$, $p{=}0.7$           & 48 & 78 & 43 \\
\bottomrule
\end{tabular}%
}
\caption{Mean frequency (\%) of the dominant vehicle across the five topics of baseline and after injection of the plausible/implausible vehicle, under varying decoding configurations. Overall, convergence is mostly stable, with expected minor variations observed: tighter top-p increases them, higher temperature decreases them.}
\label{tab:decoding-freq}
\end{table}

\begin{table}[h]
\centering
\resizebox{\columnwidth}{!}{%
\begin{tabular}{lcccc}
\toprule
Topic & \textbf{Baseline} & $t{=}1.0,p{=}1.0$ & $t{=}1.2,p{=}0.9$ & $t{=}1.0,p{=}0.7$ \\
\midrule
Time      & river      & river      & river      & river \\
Nostalgia & photograph & photograph & photograph & photograph \\
Doubt     & shadow     & shadow     & shadow     & shadow \\
Language  & map        & map        & map        & map \\
Hope      & compass    & compass    & compass    & light \\
\bottomrule
\end{tabular}%
}
\caption{Dominant vehicle per topic (baseline decoding) across configurations. Unchanged in 19/20 cells.}
\label{tab:decoding-vehicles}
\end{table}

\section{Base Models Additional Results}
\label{base_more}
\subsection{Vehicle Distribution}
\label{base}
\begin{figure*}
    \centering
\includegraphics[width=\linewidth]{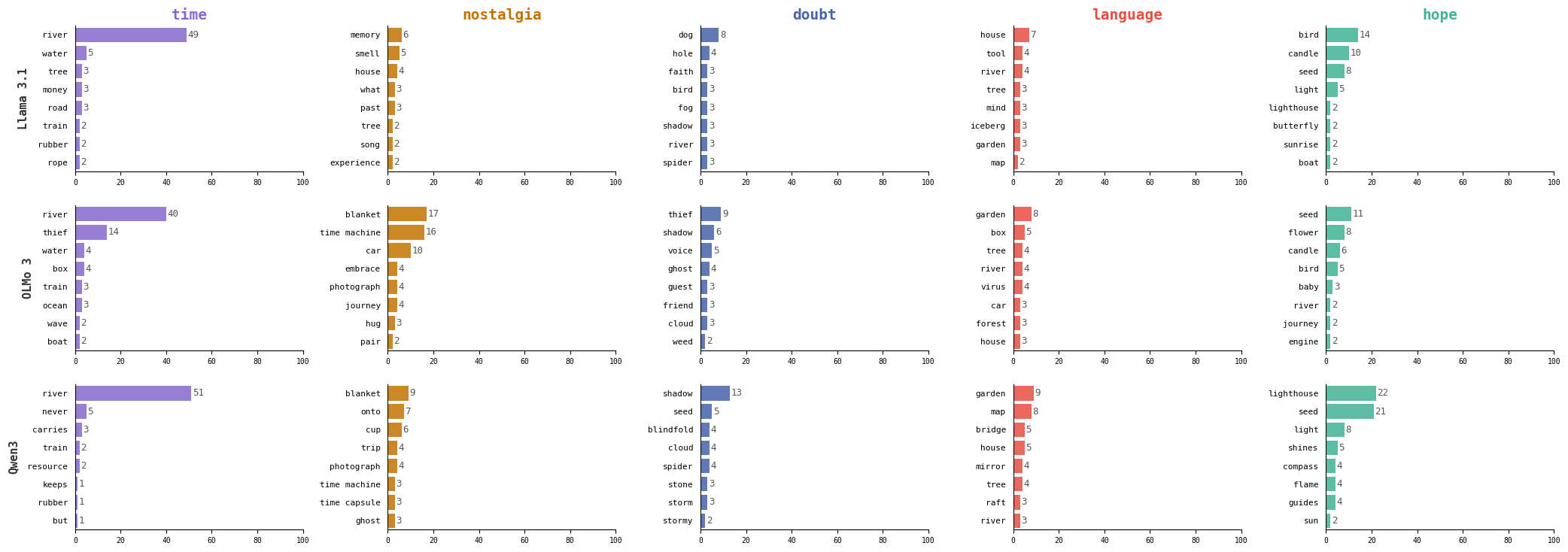}
    \caption{Full vehicle distribution under the \textbf{basic} prompting condition. Although some convergence is observed for the time metaphor, overall vehicle distributions are notably more spread out, with weaker or absent dominant vehicles across most topics. Basic prompting produces metaphors that do not follow the [topic] is [vehicle] format, making the vehicle hard to locate systematically. For uniformity, we apply the same regex function to extract the vehicle as in the other prompting conditions (and SFT experiments); as a result, the dominant vehicle is sometimes not clearly identified. We read this messiness as a feature of basic prompting: outputs are more diverse.}
    \label{fig:dist_basic}
\end{figure*}

\begin{figure*}
    \centering
    \includegraphics[width=\linewidth]{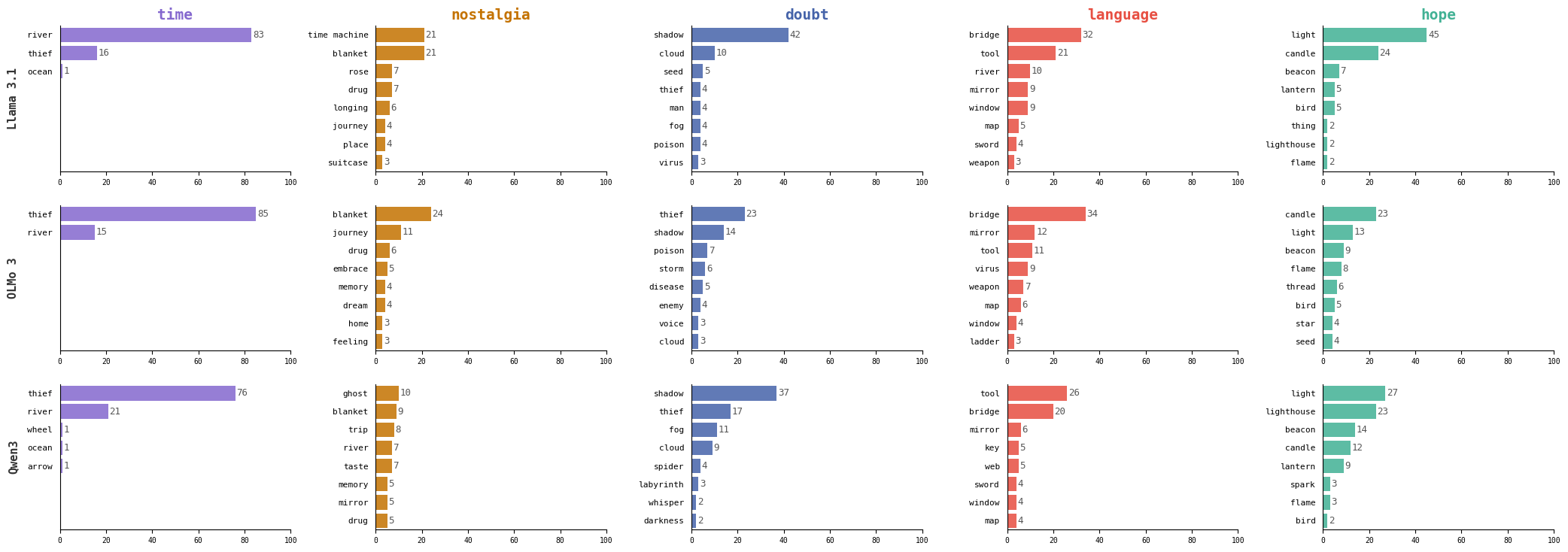}
    \caption{Full vehicle distribution under the \textbf{few-shot} prompting condition. Convergence is stronger than the basic prompting condition, but slightly lower than assistant prompting.}
    \label{fig:dist_fewshot}
\end{figure*}

\begin{figure*}
    \centering
    \includegraphics[width=\linewidth]{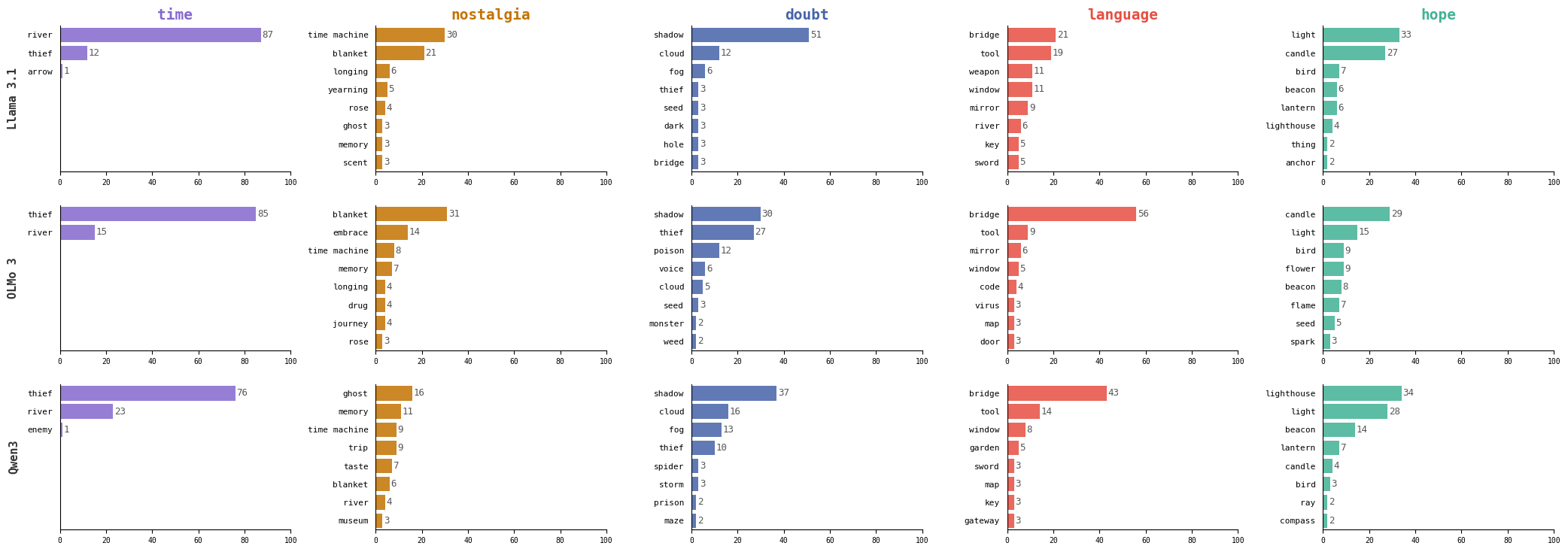}
    \caption{Full vehicle distribution under the \textbf{assistant} prompting condition across Llama 3.1 8B, OLMo 3 7B, and Qwen3 8B. Strongest convergence overall; dominant vehicles emerge across most topics and models.}
    \label{fig:dist_assist}
\end{figure*}

Full vehicle distributions across all topics, models, and prompting conditions are shown in \autoref{fig:dist_basic}, \autoref{fig:dist_fewshot}, \autoref{fig:dist_assist}. In \autoref{tab:basic-nostalgia}, \autoref{tab:fewshot-nostalgia}, \autoref{tab:assistant-nostalgia}, we provide examples of metaphors generated for the topic of nostalgia under all prompting conditions. Under the basic prompting condition, the outputs are messier and harder to parse for our regex-based parsing function.

\subsection{Few-Shot Theme Shifts Vehicle Identity}
\label{app_fewshot}

The three few-shot examples are fixed across all topics and models. To test whether the choice of examples influences vehicle distribution, we run an additional condition using nature-themed examples. The original and nature-themed prompts are shown below; the completion target \texttt{[topic] is} is appended for each topic.

\begin{small}

\noindent\textbf{Original examples:}

\begin{tcolorbox}[fontupper=\ttfamily\small, boxrule=0.4pt, colback=gray!5]
Here's a list of metaphors:\\
Life is a journey with no final destination.\\
Love is a battlefield where nobody wins.\\
Anger is a fire that consumes its host.\\
{[topic] is \textbf{...}}
\end{tcolorbox}

\noindent\textbf{Nature-themed examples:}

\begin{tcolorbox}[fontupper=\ttfamily\small, boxrule=0.4pt, colback=gray!5]
Here's a list of metaphors:\\
Life is a tide that pulls in and out without asking.\\
Love is a storm that passes and leaves the air clean.\\
Anger is a drought that hardens the ground against growth.\\
{[topic] is \textbf{...}}
\end{tcolorbox}
\end{small}

\begin{figure*}[h]
    \centering
    \includegraphics[width=\linewidth]{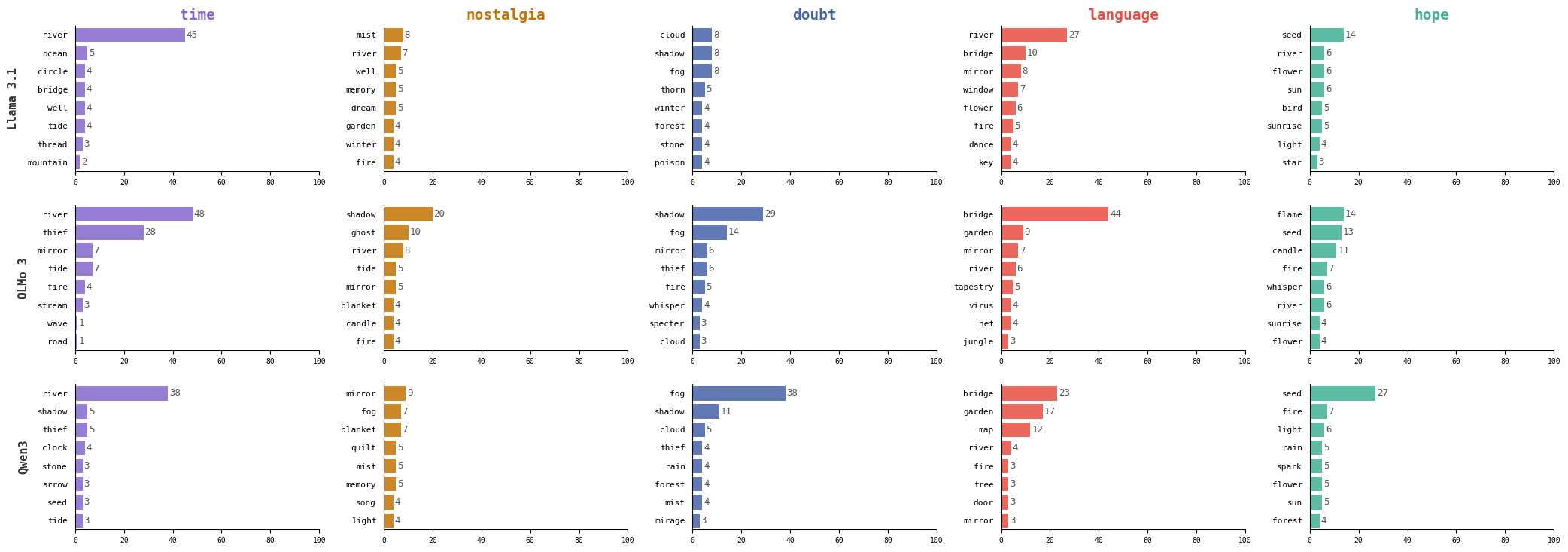}
    \caption{Full vehicle distribution under the \textbf{nature-themed few-shot} condition. Nature-related vehicles (seed, cloud, mist, fog, garden) emerge across topics.}
    \label{fig:nature_fewshot}
\end{figure*}

As shown in \autoref{fig:nature_fewshot} and \autoref{tab:fewshot-nature-nostalgia}, nature-themed examples shift vehicle distributions toward nature-related vehicles, suggesting that base models learn multiple latent preferences and that few-shot context can influence which one surfaces. \\

\subsection{Few-Shot Popularity Modulates Convergence}
\label{app_fewshot_common}

\begin{table}[h]
\centering
\resizebox{\columnwidth}{!}{%
\begin{tabular}{lccc}
\toprule
Few-shot condition & \% river & \% thief & \% other \\
\midrule
Common (rollercoaster/drug/storm) & 82 & 17 & 1 \\
Medium (library/garden/kettle)    & 27 & 2  & 71 \\
Mixed (journey/garden/microwave)  & 64 & 15 & 21 \\
\bottomrule
\end{tabular}%
}
\caption{Frequency of dominant vehicles under few-shot examples of varying popularity, on the time metaphor. We observe that no vehicle used in the few-shot examples itself becomes a converging vehicle in the tested metaphors. Medium \emph{other} cluster: 26 distinct vehicles, largest at 16\%, no alternative cluster.}
\label{tab:fewshot-acceptability}
\end{table}

In \autoref{tab:fewshot-acceptability}, we test how the choice of varying popularity or commonness can modulate convergence. Common examples are well-known, established metaphors; we use a set distinct from the ones presented in the paper to verify that results are not tied to the specific examples chosen. Medium examples are coherent but less conventional metaphors, and the mixed set combines one common, one medium, and one implausible-style example. We find that common examples produce more converged outputs, whereas examples of medium acceptability produce more diverse outputs. On the other hand, we find that a single common example (mixed) can be enough to induce convergence.

\clearpage

\begin{table*}[h]
\centering
\small
\setlength{\tabcolsep}{5pt}
\begin{tabular}{p{5cm} p{5cm} p{5cm}}
\toprule
\textbf{Llama 3.1} & \textbf{OLMo 3} & \textbf{Qwen3} \\
\midrule
\textit{a form of schizophrenia.} & \textit{a ghostly waltz through memory, with the past leading us gently\ldots} & \textit{the warm bath of memory, the gentle hush of the old jazz record\ldots} \\[4pt]
\textit{the smell of vanilla when you enter the kitchen of a grandmother\ldots} & \textit{like a forgotten library, where the walls whisper stories of past joy\ldots} & \textit{the bittersweet melody of a forgotten tune, echoing in the quiet corners\ldots} \\[4pt]
\textit{to memory what a kaleidoscope is to vision.} & \textit{like a moth to a flame, drawn in by the sweet scent of past memories\ldots} & \textit{the ghost of the past, that occasionally visits you\ldots} \\[4pt]
\textit{the ache you get when your bones remember the body you wore in the past\ldots} & \textit{like a rusty old bicycle you've left in the shed for years\ldots} & \textit{the comfort of an old shoe that always fits perfectly\ldots} \\[4pt]
\textit{an addiction to an earlier, simpler time that never really existed.} & \textit{like a broken clock, it's always telling the wrong time, but it's the only one\ldots} & \textit{the echo of a melody from a bygone era, playing softly in the depths\ldots} \\
\bottomrule
\end{tabular}
\caption{\textbf{Basic} prompting for \textit{nostalgia}. Unlike few-shot and assistant conditions, outputs vary widely in syntactic structure and do not consistently follow a \textit{nostalgia is a [vehicle]} template, making regex-based vehicle extraction unreliable.}
\label{tab:basic-nostalgia}
\end{table*}

\vspace{20mm}
% ── TABLE 2: FEW-SHOT PROMPTING ───────────────────────────────────────────────

\begin{table*}[h]
\centering
\small
\setlength{\tabcolsep}{5pt}
\begin{tabular}{p{5cm} p{5cm} p{5cm}}
\toprule
\textbf{Llama 3.1} & \textbf{OLMo 3} & \textbf{Qwen3} \\
\midrule
\textit{a time machine that takes you back to your past.} & \textit{a time machine to the past.} & \textit{a time machine that takes you back to the past.} \\[4pt]
\textit{a warm blanket that covers up the past.} & \textit{a warm blanket on a cold night.} & \textit{a warm blanket on a cold day.} \\[4pt]
\textit{a river that runs through time.} & \textit{a river that flows back to its source.} & \textit{a river that flows backward in time.} \\[4pt]
\textit{a drug that numbs the pain.} & \textit{a drug that makes you feel high.} & \textit{a drug that makes us yearn for a time that never existed.} \\[4pt]
\textit{a sweet poison that lingers in the heart.} & \textit{a bittersweet perfume of the past.} & \textit{a bittersweet longing for the past.} \\
\bottomrule
\end{tabular}
\caption{\textbf{Few-shot} prompting for \textit{nostalgia}. Outputs follow a \textit{nostalgia is a [vehicle]} structure, with vehicles beginning to cluster around shared semantic themes (time machine, warm blanket, drug, rose).}
\label{tab:fewshot-nostalgia}
\end{table*}

\vspace{20mm}
% ── TABLE 3: ASSISTANT PROMPTING ─────────────────────────────────────────────

\begin{table*}[h]
\centering
\small
\setlength{\tabcolsep}{5pt}
\begin{tabular}{p{5cm} p{5cm} p{5cm}}
\toprule
\textbf{Llama 3.1} & \textbf{OLMo 3} & \textbf{Qwen3} \\
\midrule
\textit{a time machine that transports you back in time.} & \textit{a time machine that only goes backwards.} & \textit{a time machine that takes you back to your childhood.} \\[4pt]
\textit{a warm blanket that keeps memories close.} & \textit{a warm blanket that keeps you warm on a cold night.} & \textit{a warm blanket on a cold night.} \\[4pt]
\textit{a ghost that haunts our past.} & \textit{a ghost that haunts the past.} & \textit{a ghost that haunts our memories.} \\[4pt]
\textit{a powerful drug that leaves one longing for the past.} & \textit{a drug that numbs the pain.} & \textit{a ghost that haunts the present.} \\[4pt]
\textit{a rose-colored lens through which memories are viewed.} & \textit{a bittersweet melody played on the heartstrings.} & \textit{a river flowing backward.} \\
\bottomrule
\end{tabular}
\caption{\textbf{Assistant} prompting for \textit{nostalgia}. Outputs follow the \textit{nostalgia is a [vehicle]} structure, with vehicles converging strongly across models around a small set of shared types (time machine, warm blanket, ghost, drug).}
\label{tab:assistant-nostalgia}
\end{table*}

\begin{table*}[h]
\centering
\small
\setlength{\tabcolsep}{5pt}
\begin{tabular}{p{5cm} p{5cm} p{5cm}}
\toprule
\textbf{Llama 3.1} & \textbf{OLMo 3} & \textbf{Qwen3} \\
\midrule
\textit{a mist that veils the landscape.} & \textit{a tide that rises in the heart with the moon.} & \textit{a lighthouse that beckons from the shores of memories.} \\[4pt]
\textit{a garden of memory that thrives on loss.} & \textit{a flower that blooms in the soil of memory.} & \textit{a garden of memories that withers with time.} \\[4pt]
\textit{a scent of a flower that blooms but once.} & \textit{a vine that wraps around the past.} & \textit{a field of flowers that blooms in the mind's eye.} \\[4pt]
\textit{a spring rain that nourishes new shoots in the earth.} & \textit{a moth attracted to the flame of memory.} & \textit{a river that flows gently yet carries the weight of the past.} \\[4pt]
\textit{a sunset that paints the past golden.} & \textit{a flame that burns brighter the farther it's been away.} & \textit{a gentle breeze that whispers of bygone days.} \\
\bottomrule
\end{tabular}
\caption{\textbf{Nature-themed few-shot} prompting for \textit{nostalgia}. Outputs retain the \textit{nostalgia is a [vehicle]} structure but shift vehicle types toward natural imagery (mist, tide, garden, flame).}
\label{tab:fewshot-nature-nostalgia}
\end{table*}

\end{document}